\documentclass[10pt,twocolumn,letterpaper]{article}

\pdfoutput=1

\usepackage{style/iccv}
\usepackage{caption}
\usepackage{subcaption}
\usepackage{adjustbox}
\usepackage{multirow}

\usepackage{times}
\usepackage{epsfig}
\usepackage{graphicx}
\usepackage{float}
\usepackage{amsmath}
\usepackage{amssymb}
\usepackage{bbm}
\usepackage{xcolor}
\usepackage{authblk}
\usepackage[accsupp]{axessibility}  % Improves PDF readability for those with disabilities.

\usepackage{amsmath,amsfonts,bm}

\newcommand{\norm}[1]{\left\lVert#1\right\rVert}

\def\figref#1{figure~\ref{#1}}
\def\eqref#1{equation~\ref{#1}}
\def\1{\bm{1}}

\def\rvq{{\mathbf{q}}}

\def\rmA{{\mathbf{A}}}

\def\rmF{{\mathbf{F}}}
\def\rmG{{\mathbf{G}}}
\def\rmH{{\mathbf{H}}}

\def\rmK{{\mathbf{K}}}

\def\rmM{{\mathbf{M}}}

\def\rmR{{\mathbf{R}}}

\def\rmT{{\mathbf{T}}}

\def\rmW{{\mathbf{W}}}

\def\mH{{\bm{H}}}
\def\mI{{\bm{I}}}

\DeclareMathAlphabet{\mathsfit}{\encodingdefault}{\sfdefault}{m}{sl}
\SetMathAlphabet{\mathsfit}{bold}{\encodingdefault}{\sfdefault}{bx}{n}

\def\sX{{\mathbb{X}}}

\usepackage[pagebackref=true,breaklinks=true,colorlinks,bookmarks=false]{hyperref}

\iccvfinalcopy % *** Uncomment this line for the final submission

\def\iccvPaperID{3985} % *** Enter the ICCV Paper ID here

\usepackage[font=small,skip=2pt]{caption}

\ificcvfinal\fi

\begin{document}

%%%%%%%%% TITLE
\title{\vspace{-9mm}BEV-Net: Assessing Social Distancing Compliance\\by Joint People Localization and Geometric Reasoning}
\author[1]{Zhirui Dai}
\author[1]{Yuepeng Jiang}
\author[1]{Yi Li}
\author[1,2]{Bo Liu}  % update Bo's affiliation
\author[3]{Antoni B. Chan}
\author[1]{Nuno Vasconcelos}
\affil[1]{Department of Electrical and Computer Engineering, UC San Diego}
\affil[2]{Wormpex AI Research}
\affil[3]{Department of Computer Science, City University of Hong Kong}
\affil[ ]{\texttt{\small \{zhdai,yuj009,yil898,boliu\}@eng.ucsd.edu, abchan@cityu.edu.hk, nvasconcelos@ucsd.edu}}

\maketitle
\ificcvfinal\thispagestyle{empty}\fi

%%%%%%%%% ABSTRACT
\begin{abstract}
Social distancing, an essential public health measure to limit the spread of contagious diseases, has gained significant attention since the outbreak of the COVID-19 pandemic. In this work, the problem of visual social distancing compliance assessment in busy public areas, with wide field-of-view cameras, is considered. A dataset of crowd scenes with people annotations under a bird's eye view (BEV) and ground truth for metric distances is introduced, and several measures for the evaluation of social distance detection systems are proposed. A multi-branch network, BEV-Net, is proposed to localize individuals in world coordinates and identify high-risk regions where social distancing is violated. BEV-Net combines detection of head and feet locations, camera pose estimation,  a differentiable homography module to map image into BEV coordinates, and geometric reasoning to produce a BEV map of the people locations in the scene. Experiments on complex crowded scenes demonstrate the power of the approach and show superior performance over baselines derived from methods in the literature. Applications of interest for public health decision makers are finally discussed. Datasets, code and pretrained models are publicly available at \href{https://github.com/daizhirui/BEVNet}{GitHub}.
\end{abstract}

%%%%%%%%% BODY TEXT
\vspace{-1em}
\section{Introduction}
\vspace{-0.5em}

% The COVID-19 pandemic\dots

Social distancing, the strategy of maintaining a safe distance between people in public spaces, has been shown to be an effective measure against the transmission of contagious pathogens, including influenza virus and coronavirus \cite{siegel2007guideline,courtemanche2020strong,hsiang2020effect}. However, the monitoring of social distancing by human observers is neither practical in many settings nor scalable. This has motivated an interest in methods to detect and count social distancing violations automatically. While non-vision-based methods are available \footnote{The \emph{DP-3T} contact tracing protocol~\cite{troncoso2020decentralized}, for example, estimates distances using Bluetooth signal on smartphones.},
they typically require users to install certain applications on their mobile devices, and are limited in precision of distance estimates.

\begin{figure}
    \centering
    \includegraphics[width=0.8\linewidth]{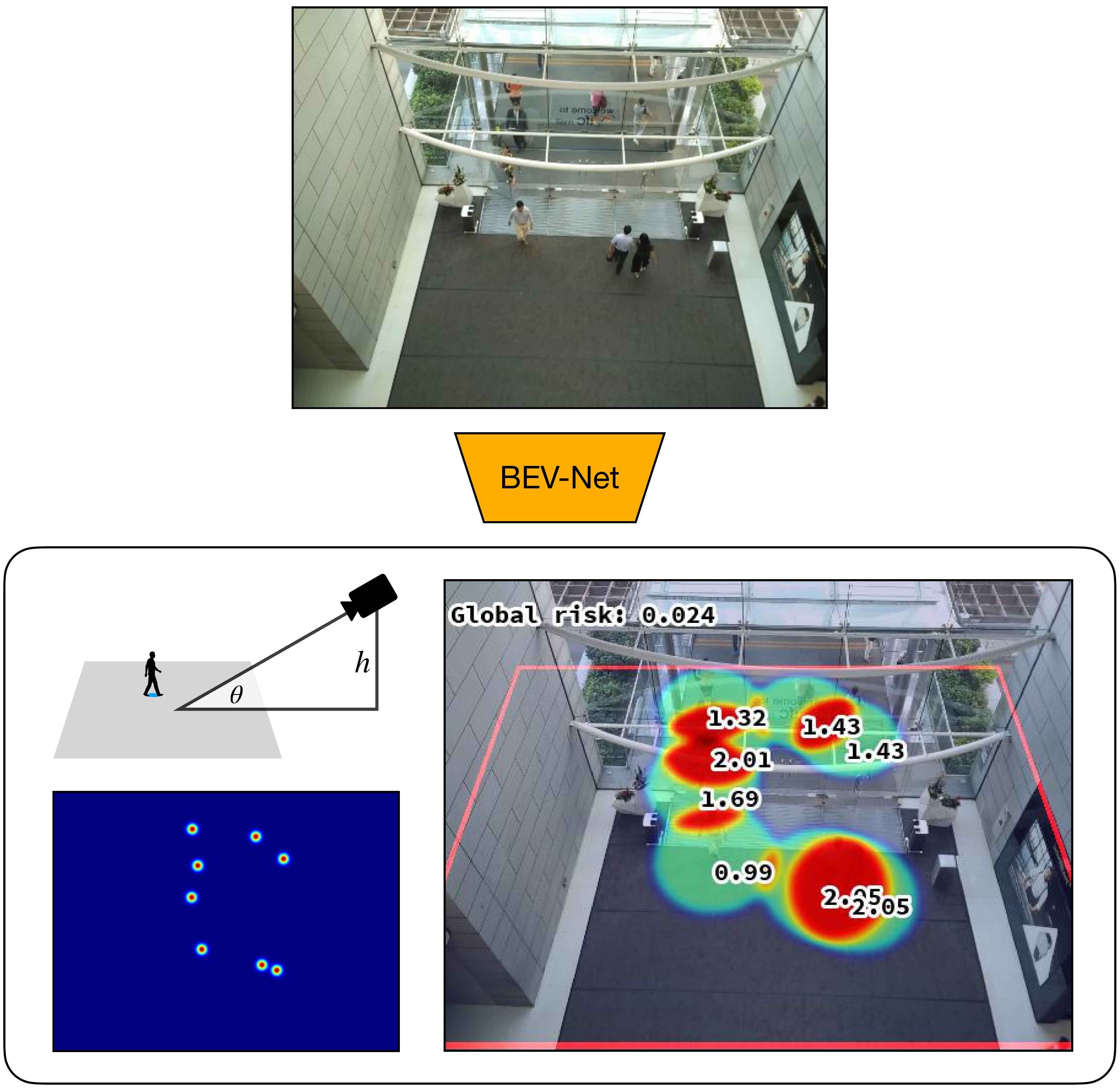}
    \caption{Set of tasks proposed for social distancing compliance assessment. Given an input image, models are expected to jointly predict camera geometry (\textbf{top left}), bird's eye view heatmap of person locations (\textbf{bottom left}), local risk map highlighting areas with high probability of infection (\textbf{right}), as well as individual and global risk level for the entire scene (annotated over the risk map).
    }
    \label{fig:teaser}
\end{figure}

Computer vision offers a viable alternative for the collection of social distance measurements. In particular, it has several advantages for the monitoring of public spaces. First, it can leverage surveillance cameras that are already available in many public locations. No expensive infrastructure changes are required.

Second, anonymization of visual data is straightforward by removing all facial identities, as the system has no access to other sensitive information of pedestrians.
This makes it much more privacy preserving than the monitoring of mobile devices, or similar approaches. On the other hand, it can produce complete statistics of social distancing violations, as there is no prerequisite on the population (e.g. smartphone with Bluetooth enabled).
While not suited for contact-tracing, these statistics can be very useful to decision makers, e.g. to enable the implementation of highly localized ``lock-downs,'' time-varying control of pedestrian access to certain areas, etc.

Computer vision has a long history of sensing humans in images. Object detection \cite{felzenszwalb2009object,girshick2014rich,girshick2015fast,ren2015faster,lin2017feature} and instance segmentation \cite{dai2016instance,he2017mask,li2017fully} recognize and localize objects by bounding boxes or pixel-wise masks. While effective for close objects and sparsely populated scenes \cite{everingham2010pascal,lin2014microsoft}, detection quality degrades substantially for far cameras and busy spaces, involving significant degrees of occlusion. In fact, on such scenes, it is even impossible to collect accurate bounding box annotations.
Social distancing measuring is more closely related to crowd-counting methods \cite{Zhang_2016_CVPR,cao2018scale,crowd_counting_context_aware}, which are trained to produce a heatmap that highlights the head location of every person in the scene. However, because these methods don't explicitly reason about scene geometry, they are unsuitable to estimate distances between individuals. Furthermore, head locations are not ideal for estimating scene distances, since heads do not lie on a shared plane in 3D, due to people height variation. Much more accurate distance estimates can usually be obtained by reconstructing the scene ground plane and measuring feet distances. This, however, presents additional challenges, due to occlusion.

In this work, we consider the problem of social distancing compliance assessment (SDCA), which aims to measure the distances between individuals in a scene and detect violations of social distancing thresholds. Based on the above observations, we argue that SDCA requires a geometry-aware approach, where the relevant extrinsic parameters of the camera are estimated and used to generate a bird's eye view (BEV) map of people locations, as illustrated in Figure~\ref{fig:teaser}. We propose a novel benchmark for SDCA, CityUHK-X-BEV, which repurposes the CityUHK-X dataset \cite{kang2017incorporating} to the SDCA problem, by adding ground-plane annotations. Specifically, for each head location in the dataset, the corresponding feet position is annotated, and mapped to the BEV, using the known intrinsic and extrinsic camera parameters. A novel set of evaluation criteria is introduced, each focusing on a different aspect of the task: \emph{Localization} metric that measures the accuracy in detecting real-world locations of people in the ground plane; \emph{local risk} metric that evaluates the capability to discover regions with high chance of infection; \emph{global risk} metric that predicts the overall risk level of captured scene. Figure~\ref{fig:teaser} shows examples of these tasks. Unlike many vision problems that localize object in the image plane, these geometry-aware criteria directly evaluate the capacity of models to make predictions in the 3D ground plane, leading to outputs that are much more informative for real-world applications that require metric information.

A multi-branch convolutional architecture, BEV-Net, is then proposed to solve the SDCA task. BEV-Net follows an encode-decoder structure, using a projective transformation module to convert convolutional feature maps from image view to BEV. The decoder is implemented with a pose regression branch that estimates camera parameters and three separate branches to predict feet, head and BEV heatmaps.
These branches are trained with individual losses, in a multi-task manner. To compensate for the height variations in the crowd, a group transformation module with spatial self-attention is used to group people by head height and independently align the feet and head feature maps of the resulting groups. Experiments show that the BEV-Net outperforms all DET and CC baselines under all proposed SDCA evaluation metrics. It is further shown, through ablation experiments, that both head and feet annotations are essential to achieve the best prediction quality.

A number of applications of potential interest for public health decision makers are then illustrated. These range from the characterization of risks for a single image, as illustrated in Figure~\ref{fig:teaser}, to global measures of scene risk, integrated over image datsets, as shown in Figures~\ref{fig:top_risk} and \ref{fig:scene_risk}. The latter can be used to identify events of unusually large risk or inform the deployment of risk mitigation measures, such as the introduction of obstacles in the scene to modify walking patterns and other crowd behaviors.

The paper makes four major contributions: First, we introduce the idea of using computer vision for joint geometric reasoning and social distancing compliance assessment on public spaces. Second, a novel benchmark for SDCA in crowd scenes, CityUHK-X-BEV, is introduced with person-level annotations in bird's eye view. Third, a multi-branch convolutional network, BEV-Net, is proposed and shown to achieve best SDCA results by learning to perform both heatmap prediction and geometric reasoning. Finally, we show promising results for several potential applications of the SDCA framework in the public health domain.

\vspace{-0.5em}
\section{Related Work}
\vspace{-0.5em}

\noindent
{\bf Object detection (DET).}
Object detection methods recognize and localize multiple classes of objects with bounding boxes. While early algorithms relied on hand-crafted visual features \cite{lowe2004distinctive,dalal2005histograms}, the introduction of CNN-based detectors \cite{girshick2014rich,girshick2015fast,ren2015faster,cai2018cascade,redmon2016you,lin2017focal} trained on large-scale image databases \cite{deng2009imagenet,lin2014microsoft} has enabled dramatic performance gains.

Existing approaches to SDCA have mostly relied on pretrained DET models, making few technical advances to their architectures. \cite{interhomines} proposed to detect social distancing violations by regressing head and feet locations from the bounding box of each detected person. While effective for sparsely populated environments, such an approach does not scale to busy spaces and distant cameras with a large field of view, as is usually the case of large public spaces. In addition, \cite{interhomines} requires external homography calculation based on markers manually placed in the scene. By contrast, the proposed BEV-Net automatically estimates camera geometry and is able to estimate social distances for crowded environments and wide field of view cameras.

\noindent
{\bf Crowd counting (CC).}
Crowd counting focuses on counting the number of people in an image. State-of-the-art methods either learn to regress head counts from the image data directly~\cite{marsden2017fully}, or predict a people density map which is then integrated to obtain the people count~\cite{Zhang_2016_CVPR,sam2017switching,cao2018scale,li2018csrnet,liu2019crowd}. Crowd scene datasets tend to be collected in public spaces and focus on busy scenes~\cite{zhang2015cross,idrees2013multi,wang2020nwpu,kang2017incorporating}, where the estimation of head locations is difficult due to small people sizes and significant occlusion. These are the scenes that we emphasize in this work, where we augment a popular crowd counting dataset with rich annotations required for SDCA.

Most CC methods are trained to produce density maps in the camera view, a simpler problem than the proposed combination of SDCA tasks. An exception is WACC~\cite{zhang2019wide}, which learns to predict ground-plane heatmap directly. The BEV-Net differs from this work in three main aspects: First, WACC requires inputs from multiple cameras, while BEV-Net is designed to work with a single camera view. Second, WACC is supervised by head annotations only, leading to inaccurate ground-plane locations due to varying head heights; BEV-Net addresses this by producing feet annotations that, unlike heads, lie on the common ground plane. Third, WACC assumes known geometry for all camera views, while BEV-Net jointly learns to predict extrinsic camera parameters.

\noindent
{\bf Geometry in computer vision.} The scene geometry recovery is a classical problem in computer vision~\cite{zissermanHartley}. This can be decomposed into the estimation of camera parameters and scene geometry, i.e., depth variability in different parts of the scene. Since the introduction of deep learning, both components have been estimated by neural networks, typically by using two dedicated branches that are trained jointly, in an end-to-end manner~\cite{laina2016deeper,zhou2017unsupervised,godard2017unsupervised,kanazawa2018end}. The scene reconstruction required by SDCA amounts to recovering the ground plane and 3D feet locations of all individuals. This is not trivial because feet locations are frequently occluded in the camera view. BEV-Net addresses this by leveraging head locations and the regularizing geometric constraint that a standing person's head and feet are co-located in BEV.

\noindent
{\bf Social behavior analysis.} Computer vision methods have been applied to modeling human behavior in public spaces. One line of work achieves this through the task of trajectory prediction~\cite{lerner2007crowds,pellegrini2009you,kitani2012activity,alahi2016social,gupta2018social}, which requires generating plausible motion paths for pedestrians in the image plane.
Early work used physics models such as Social Force~\cite{helbing1995social} to account for interaction between humans~\cite{mehran2009abnormal,pellegrini2010improving,yamaguchi2011you,robicquet2016learning}; more recently, neural network modules were used to capture such dependencies between agents, e.g. with recurrent networks~\cite{alahi2016social,bartoli2018context,sadeghian2019sophie} or graph convolutional networks~\cite{mohamed2020social,sun2020recursive}. Other tasks for social behavior modeling have also been explored, including early action detection~\cite{ryoo2011human,koppula2015anticipating,ma2016learning} and group activity recognition~\cite{choi2011learning,lan2012social,ibrahim2016hierarchical}. All of the above tasks require temporal modeling on video data and semantic understanding of human activities; this work instead focuses on sensing the spatial locations of people, which can be efficiently recovered from individual image frames.

\section{Social Distancing Compliance Assessment}
\vspace{-0.5em}

To the best of our knowledge, no prior work has attempted to evaluate the quality of visual SDCA in busy public spaces and large field-of-view scenes. In this section, we propose a new dataset for this task.

\begin{figure}[t]
    \centering
    \includegraphics[width=0.85\linewidth]{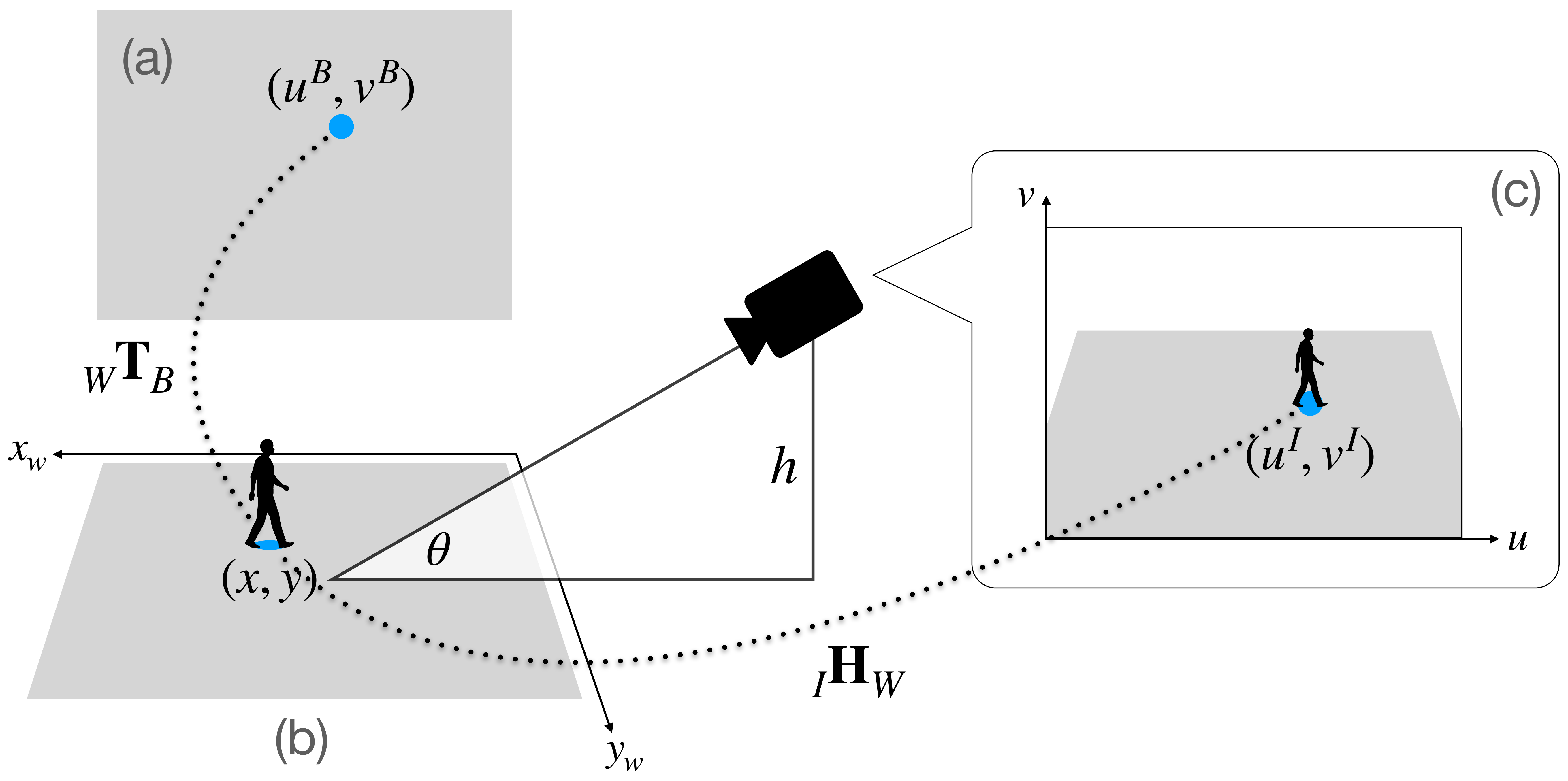}
    \caption{Homography between BEV, IV, and world coordinates. \textbf{(a)} BEV map with person location $(u^B, v^B)$; \textbf{(b)} world model with camera at height $h$, pitch angle $\theta$ and person at position $(x, y)$ on the ground plane; \textbf{(c)} image captured by camera with feet detected at pixel location $(u^I, v^I)$. Coordinates are converted across views through homography matrices ${_W\rmT_B}$ and ${_I\rmH_W}$; gray area denotes region of interest.}
    \label{fig:homography}
\end{figure}

\vspace{-0.5em}
\subsection{Motivation}
\vspace{-0.5em}

A dataset for SDCA should satisfy various requirements. First, it should contain a wide range of scenes with varying people densities.
Second, it should include ground-truth locations for the people in the scene, either in 3D world coordinates or in the form of a BEV heatmap. Optionally, it could provide ground-truth for the intrinsic and extrinsic camera parameters, allowing direct camera pose supervision during training and facilitating the recovery of ground plane and homography between camera view and BEV.

DET datasets~\cite{dollar2011pedestrian,lin2014microsoft,zhang2017citypersons} are not suitable for SDCA since the number of people per image tends to be low. CC datasets are more relevant, as they contain abundant examples of people gathering in clusters---often within 1 to 2 meters, the range of droplet transmission \cite{wells1934air,fernstrom2013aerobiology,kutter2018transmission}---and the scene/camera configurations are most suitable for monitoring social distances in public spaces. However, geometric meta-information is unavailable in most CC datasets.

CityUHK-X~\cite{kang2017incorporating} is an exception, providing extrinsic camera parameters, including height and pitch angle, which make it a potential SDCA benchmark. Nevertheless, it has limitations. Like other CC datasets, it only provides image annotations for each \emph{head} in the scene. Even with known camera parameters, image head locations aren't enough to recover locations in world coordinates, as each individual's height is variable and unknown. Hence, additional annotations are needed for the \emph{feet} locations of each person in the image, as well as head-feet correspondences.

\vspace{-0.5em}
\subsection{Annotation}
\vspace{-0.5em}

The Amazon MTurk platform was used to collect feet annotations for CityUHK-X with correspondences between feet and heads. Given a crop of the scene image with annotated head location, MTurk workers were asked to annotate the center point between both feet, and to verify if the feet were clearly visible (feet location is precise) or occluded (feet location is estimated). 87,746 feet locations were annotated in the 2,982 scene images in CityUHK-X. Among them, 63,669 (72.6\%) were clearly visible and 24,077 (27.4\%) occluded. Detailed description of the annotation procedure can be found in the supplemental.

\vspace{-0.5em}
\subsection{Ground-truth Generation} \label{sect:gt_map}
\vspace{-0.5em}

\begin{figure}[t]
    \centering
    \includegraphics[width=0.8\linewidth]{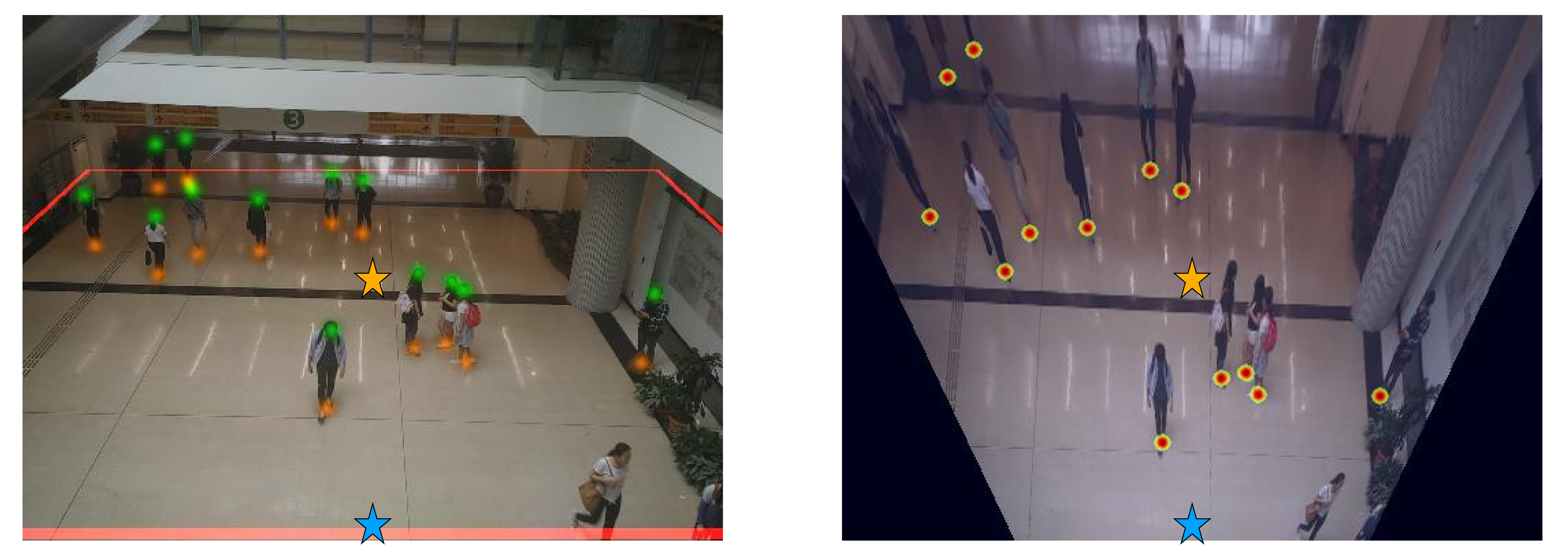}
    \caption{Example ground-truth heatmaps overlaid on the input image. \textbf{Left}: IV maps, with {\color{green} heads} in green and {\color{red} feet} in red. Region of interest is bounded by a red perspective box. \textbf{Right}: BEV location map. {\color{orange} Centers} and {\color{cyan} bottom centers} of heatmaps are aligned.}
    \label{fig:ground_truth_example}
\end{figure}

As illustrated in Figure \ref{fig:ground_truth_example}, ground-truth is provided in the form of three density maps: feet and head maps in image view (IV maps), and person location map in BEV coordinates (BEV map).

\noindent
{\bf Image view maps.}
Following \cite{Zhang_2016_CVPR}, the ground-truth head and feet locations in image view are represented with heatmaps $\rmM_{\textrm{head}}$ and $\rmM_{\textrm{feet}}$ composed by mixture of Gaussians. Each Gaussian from the head (feet) heatmap is centered at the annotated head (feet) location of a person in the image, with a fixed standard deviation $\sigma=5$ pixels.

\noindent
{\bf Bird's eye view map.}
Given the image coordinates of annotated keypoints and camera geometry, the BEV map $\rmM_{\textrm{BEV}}$ is generated using a homography.
This is achieved by projecting the keypoints to world coordinates, choosing a suitable region of interest within the ground plane, then resampling into a fixed-size BEV heatmap, as illustrated in Figure~\ref{fig:homography}.
First, the world coordinate frame is defined by setting the ground plane to $z = 0$ and the camera sensor to $(0, 0, h)$, where $h$ denotes camera height above ground. The camera's yaw angle is zero in this setting. Further assuming that the camera has pitch angle $\theta$, and zero roll (the roll angle could be zero by transforming the input image properly),
the transformation between world coordinates $(x,y)$ in the ground plane and image coordinates $(u^I, v^I)$ is then given by the homography (derivations in supplemental)
\begin{equation}
    \left[\begin{matrix}u^I & v^I & 1\end{matrix}\right]^\top = {_I\rmH_W} \left[\begin{matrix}x & y & 1\end{matrix}\right]^\top \label{eq:imgh},
\end{equation}
where
$_I\rmH_W$ is constructed as a function of $h, \theta$ and intrinsic camera parameters such as focal lengths $(f_u, f_v)$.

Second, to define a proper region of interest (RoI) on the ground plane, we require that the center $(u_c^B, v_c^B)$ of BEV map be projected to the image center $(u_c^I, v_c^I)$, and the bottom-center pixel in BEV be aligned with the bottom-center image pixel, as indicated in \ref{fig:ground_truth_example}.
The scale factor $s$, measuring distance in meters spanned per BEV pixel, is then given by
% \vspace{-0.8em}
\begin{align}
    s = 2(x_c - x_{bc})/H = {h} / \left[{\beta(v_c^I\alpha+f_v\beta)}\right]
    \label{eq:bev_scale}
%    s = \frac{x_c - x_{bc}}{H/2} = \frac{h}{\beta(v_c^I\alpha+f_v\beta)}
%    \label{eq:bev_scale}
\end{align}
where $H$ is the height of the BEV map.
The transformation from BEV map coordinates $(u^B, v^B)$ to world coordinates in the ground plane is then given by the homography
\begin{align}
    \left[\begin{matrix}x & y & 1\end{matrix}\right]^\top = {_W\rmT_B} \left[\begin{matrix}u^B & v^B & 1\end{matrix}\right]^\top,
    \label{eq:BEVh}
\end{align}
where
$_W\rmT_B$ is constructed to align the BEV map with image center, and rescale to $s$ meters per BEV pixel.
%where $(x_c, y_c)$ is the world coordinate of the image center $(u_c^I, v_c^I)$.
Finally, the homography between image and BEV map coordinates is obtained by combining (\ref{eq:imgh}) and (\ref{eq:BEVh}) into
\vspace{-0.5em}
\begin{align}
    \left[\begin{matrix}u^I & v^I & 1\end{matrix}\right]^\top = {_I\rmH_B} \left[\begin{matrix}u^B & v^B & 1\end{matrix}\right]^\top, \label{eq:bev_transform}
\end{align}
\vspace{-.5em}
where
\begin{equation}
    {_I\rmH_B} = {_I\rmH_W}\ {_W\rmT_B}.
    \label{eq:iHB}
\end{equation}

Given a set of feet locations $\{\rvq_j\}_{j=1}^d$ in image $\mI$, the corresponding locations in BEV coordinates are given by $\{{_I\rmH_B}^{-1}\rvq_j\}_{j=1}^d$.
Similar to the IV maps, the BEV map $\rmM_{\textrm{BEV}}$ is generated using a Gaussian kernel with $\sigma = 5$ px.
Example heatmaps are shown in \figref{fig:ground_truth_example}.

\subsection{Evaluation protocol} \label{sect:eval_protocol}
\vspace{-0.5em}

We propose two types of criteria for evaluating the quality of predicted BEV heatmaps for SDCA.

\noindent
{\bf Localization error.}
Models are required to identify BEV locations in real-world distance units (e.g. meters or feet), for all individuals in the scene.
Given a set of predicted locations $\hat{\sX} = \{\hat{\bf x}_i\}_{i=1}^M$ and a set of ground-truth locations $\sX = \{{\bf x}_i\}_{i=1}^N$, the Chamfer distance~\cite{barrow1977parametric}
\begin{equation}
\small
D(\hat{\sX}, {\sX}) = \frac{1}{M} \sum_{i=1}^M \min_j \norm{\hat{\bf x}_i - {\bf x}_j} + \frac{1}{N} \sum_{j=1}^N \min_i \norm{\hat{\bf x}_i - {\bf x}_j}
\end{equation}
is used to evaluate localization error in terms of real-world distances. Predicted locations $\hat{\sX}$ are determined from the BEV heatmap, using non-maximum suppression of size $5 \times 5$ pixels, followed by pixel-wise thresholding at the heatmap value $10^{-3}$. The non-zero entries of the post-processed BEV map are then extracted and converted to world coordinates using (\ref{eq:BEVh}). We also evaluate the normalized chamfer distance $D_n(\hat{\sX}, {\sX}) = \frac{D(\hat{\sX}, {\sX})}{2 d_0}$, which measures the localization error as a percent of the safe distance threshold. A minimum safe distance of $d_0 = 1.5$m is used, but can be adjusted per public health guidelines.

\noindent
{\bf Risk estimation error.} Models are expected to measure compliance with social distancing by estimating risk levels in the scene, either locally or globally. \emph{Local} risk levels are represented as a heatmap $\rmR$ on the ground plane, with greater values indicating locations with higher risk of infection. Risk is estimated from the BEV localization map $\rmM_{\textrm{BEV}}$ of sect.~\ref{sect:gt_map} by applying a scale-adaptive kernel $\rmK$ determined by a chosen infection risk model.
\begin{equation}
    \rmR(u, v) = \rmM_{\textrm{BEV}}(u, v) * \rmK\left(u, v\right).
\end{equation}
When the infection risk is defined as simply the number of people within the safe distance to a person, $\rmK$ is a disk-shaped kernel of radius $r = d_0 / s$, where $s$ is the scale parameter of \eqref{eq:bev_scale}. Therefore, $\rmR(u, v)$ represents the count of people within radius $r$ of location $(u, v)$. Evaluating the accuracy of the risk map by comparing pixel-wise values can be sensitive to overcrowded areas and fail to capture borderline cases. Instead, we pose local risk estimation as a segmentation problem: For a given image, the network outputs a binary mask by thresholding the risk heatmap, with positive regions indicating areas where transmission is likely to occur; the prediction quality is evaluated by intersection-over-union (IoU) \wrt ground-truth mask.

\emph{Global} risk levels are defined by counting occurrences of social distancing violation---the total number of people that fail to maintain a minimum distance $d_0$ from one another---and normalizing by the area covered by the BEV map. This is estimated by multiplying the BEV heatmap with the binary risk mask, then integrating over the space
    \vspace{-4pt}
\begin{equation}
    R_g(r_0) = \frac{1}{s^2 HW} \sum_{\substack{u, v \\ R(u, v) \ge r_0}} M_{\textrm{BEV}}(u, v)
    \vspace{-8pt}
\end{equation}
where $r_0$ is a threshold on the acceptable risk level. Risk above $r_0$ is considered unsafe, indicating the possibilty of infection due to violation of social distancing recommendations. Global risk error is measured by mean squared error (MSE) between estimated and ground-truth risks.

Figure \ref{fig:teaser} shows sample outputs for the tasks described above. The BEV heatmap $\rmM_{\textrm{BEV}}$ relies on accurate detection of each individual, while local and global risk estimates are more robust to minor localization errors, as the influence of each person is spread out in the ground plane.This diverse set of criteria assures that model outputs perform well with respect to both metrics of interest for computer vision (localization) and public health practitioners (risk levels).

\vspace{-0.5em}
\section{BEV-Net}
\vspace{-0.5em}

In this section we present BEV-Net, a unified framework for the solution of crowd counting, camera pose estimation and social distancing compliance assessment.

\vspace{-0.5em}
\subsection{Multi-branch Encoder-Decoder}
\vspace{-0.5em}

The design of BEV-Net is based on the encoder-decoder architecture commonly used in CC models \cite{cao2018scale,crowd_counting_context_aware}. However, we have found that directly training an encoder-decoder to generate BEV heatmaps leads to poor results, since a fully convolutional architecture has difficulty modelling the large and non-uniform displacements that exist between a pixel location in the input image and the corresponding location in the BEV map.

\begin{figure}[t]
    \centering
    \includegraphics[width=0.75\linewidth]{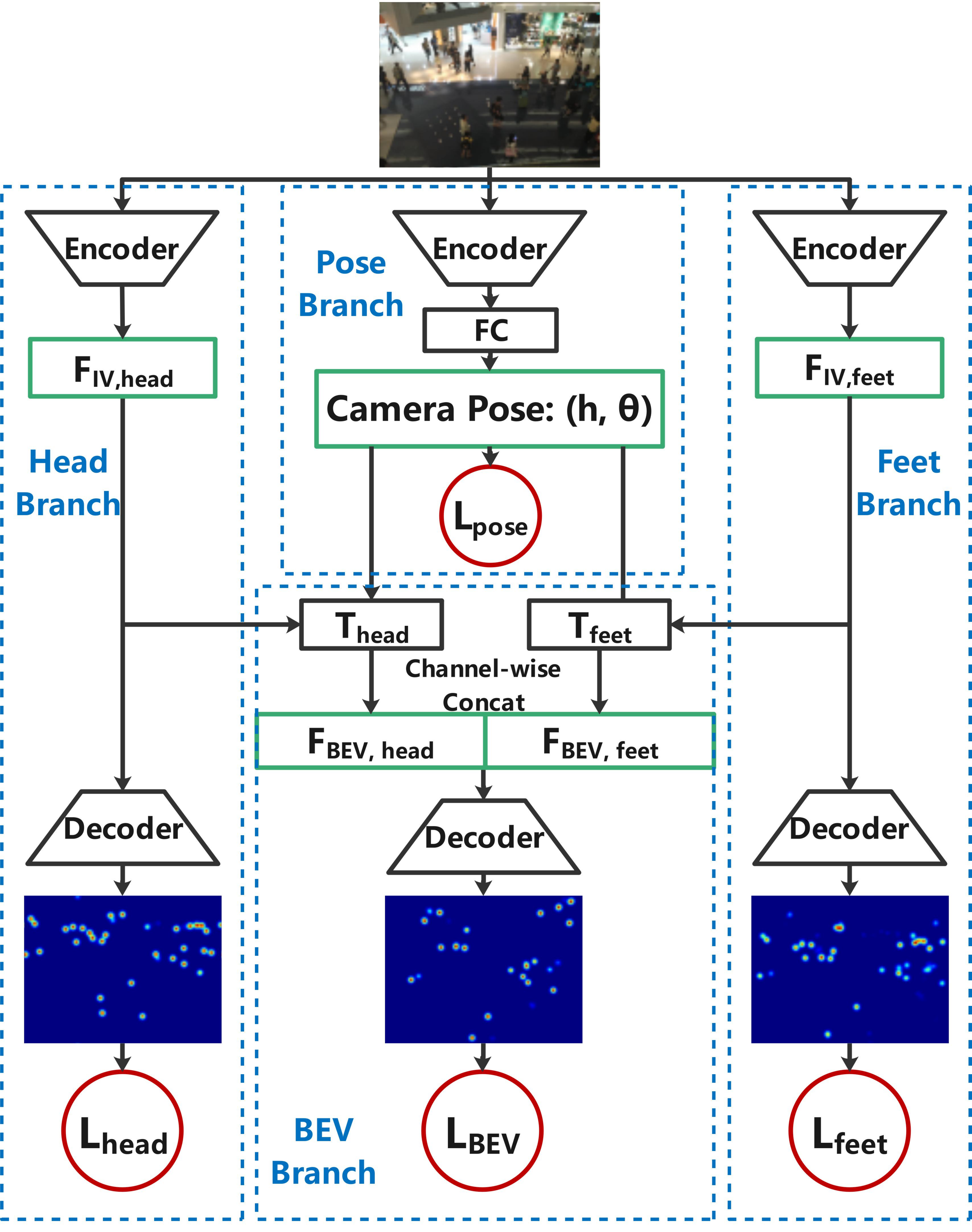}
    \caption{The BEV-Net architecture is a multi-branch model that jointly performs geometric reasoning (\textbf{pose} branch), image-view keypoint localization (\textbf{head} \& \textbf{feet} branch), and bird's eye view person localization (\textbf{BEV} branch).}
    \label{fig:net_arch}
\end{figure}

BEV-Net addresses this problem through the multi-branch architecture of \figref{fig:net_arch}. The head and feet branches are trained to predict heatmaps for head and feet in the image view (IV), respectively. Standard convolutional encoder and decoder layers suffice to implement these branches, as the input image and the IV heatmaps are aligned. For SDCA, these heatmaps are not of interest per se, since they contain no metric information. However, the addition of the two branches enables supervision for head and feat locations, which is critical to let the network select which image features to pay attention to. In this sense, they can be seen as a top-down attention mechanism.

All metric information is recovered by the central BEV branch. This branch has two stages. The first is a pose regression network that runs in parallel with head and feet branches, enabling geometric reasoning by learning to predict the height $h$ and pitch angle $\theta$ of the camera. This is implemented with a CNN feature extractor, followed by layers of MLP, and supervised by a pose-estimation loss. The second stage uses the camera parameters to rectify the IV head and feet feature maps, denoted $\rmF_\text{IV, head}$ and $\rmF_\text{IV, feet}$, into BEV coordinates. Given the predicted camera pose $(\hat{h}, \hat{\theta})$, the IV feature maps are first aligned in BEV space through projective transformation $T_\text{head}$ and $T_\text{feet}$ (details in section \ref{sect:bev_transform} and \ref{sect:group_transform}). This produces a pair of feature maps $\rmF_\text{BEV, head}$ and $\rmF_\text{BEV, feet}$ in BEV, which are then concatenated along channel dimension and fed to the BEV decoder, eventually producing the predicted BEV map.

\vspace{-0.5em}
\subsection{BEV-Transform: Feature-level Homography} \label{sect:bev_transform}
\vspace{-0.5em}

The projective transformation between IV and BEV (see \figref{fig:ground_truth_example}) creates a spatially varying displacement between IV and BEV feature map locations. This makes it difficult to predict the BEV map from the IV feature maps, since the convolution operation is not naturally suited to model spatially varying displacements. Inspired by \cite{jaderberg2016spatial}, we address this problem by designing a differentiable BEV Homography transformation module based on (\ref{eq:bev_transform}), called BEV-Transform, to perform feature-level homography mapping.

Given the predicted camera pose $(\hat{h}, \hat{\theta})$, for a plane at height $h_0$, BEV-Transform calculates the homography transformation $\hat{_I\rmH_B} = H(\hat{h} - h_0, \hat{\theta})$ that maps BEV map grid $\rmG_B$ into the IV map sample grid $\rmG_I$:
\begin{align}
    \left[\begin{matrix}u^I & v^I & 1\end{matrix}\right]^\top = \hat{_I\rmH_B} \left[\begin{matrix}u^B & v^B & 1\end{matrix}\right]^\top, \label{eq:grid_transform}
\end{align}
where $(u^B, v^B)$ and $(u^I, v^I)$ are coordinates in $\rmG_B$ and $\rmG_I$ respectively.

Note that $h_0$ is different for head and feet planes. Hence, two matrices $(\hat{_I\rmH_B})_i, i \in \{\text{head, feet}\}$ are needed to transform the feature maps of head and feet, respectively, in order to align them in BEV. In practice, more matrices are used as revealed in Section~\ref{sect:group_transform}.

Given these matrices, (\ref{eq:grid_transform}) is then used to transform the feature maps from image to BEV coordinates, using
\vspace{-0.5em}
\begin{align}
    (\rmF_{\text{BEV}, i})_{u^B, v^B} = (\rmF_{\text{IV}, i})_{u^I, v^I},\quad i \in \{\text{head, feet}\}
\end{align}

As in \cite{jaderberg2016spatial}, this is implemented with a differentiable bi-linear interpolation layer.
It should be noted that when feature maps are internally resized by the network, the coordinates $(u_c^I, v_c^I)$ of image center and focal lengths $(f_u, f_v)$ are scaled proportionally to match the map size.

\vspace{-0.5em}
\subsection{Group BEV-Transform with Spatial Attention} \label{sect:group_transform}
\vspace{-0.5em}

The determination of feet and head plane heights has different complexity. For feet, the height is determined trivially as $h_0=0$. For heads, however, since people's heights are different, the head annotations are not in the same horizontal plane in the world frame. One possibility would be to simply ignore head locations. However, the co-location of the vertical projections of head and feet is a strong regularization constraint for camera pose estimation.

\begin{figure}[t]
    \centering
    \includegraphics[width=0.85\linewidth]{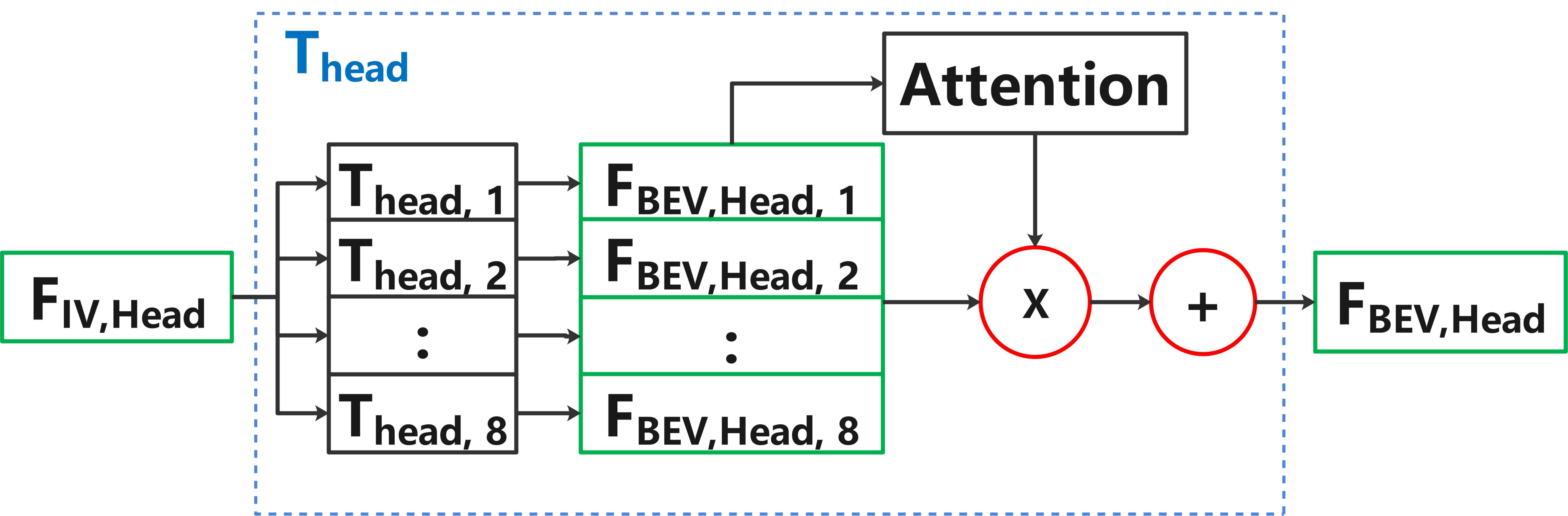}
    \caption{Architecture of Grouping BEV transform modules and spatial attention mechanism}
    \label{fig:multi_attn}
\end{figure}

To take advantage of this, BEV-Net relies on
%a set of {\it approximate\/} people heights, implemented with
a set of head planes that quantify the range of person heights. To cover both adult and child heights, planes are placed at heights 1.1m, 1.2m, ..., 1.8m from the ground.
People in different regions of the image are then automatically assigned to different height planes by a self-attention mechanism, as illustrated in \figref{fig:multi_attn}. The IV head feature maps $(\rmF_{\text{IV}})_{head}$ are first mapped by the homographies $(\hat{_I\rmH_B})_{head,i}, i \in \{1,\ldots,8\}$, associated with the different head heights, into a set of BEV head feature maps $\{(\rmF_{\text{BEV}})_{head,i}\}$. An attention branch consist of three convolutional layers is then used to compute a weight map $\rmW_i$ per BEV head feature map. The set of weight maps $\{\rmW_i\}$ are then normalized by a 2D softmax layer, notated as $\{\tilde{\rmW}_i\}$. The resulting spatially varying weighted combination of the feature maps $\sum_{i=1}^8 (\rmF_{\text{BEV}})_{head,i} * \tilde{\rmW}_i$ is finally used as the BEV head feature map $(\rmF_{\text{BEV}})_{head}$.

\vspace{-0.5em}
\subsection{Multi-task Loss}
\vspace{-0.5em}

As shown in \figref{fig:net_arch}, BEV-Net is trained with four loss functions. The loss functions for the head, feet and BEV branches are all MSE losses:
\vspace{-0.5em}
\begin{align}
    L_i = \text{MSE}(\hat{\rmM_i}, m \rmM_i) + \alpha \text{MSE}(g(\frac{\hat{\rmM_i}}{m}), g(\rmM_i))
\end{align}
where $\alpha = 1.25\times 10^{-6}, \quad i \in \{\text{head, feet, BEV}\},\ g(\rmA) = \sum_{u,v} A_{u,v}$ gives the global people counting. We amplify ground-truth heatmaps by a factor of $m = 100$ for better convergence, following practices in~\cite{Zhang_2016_CVPR}.
The pose loss combines two MSE losses, for camera height and pitch angle:
\vspace{-0.5em}
\begin{align}
    L_{\text{pose}} = \lambda_{\text{angle}}(\hat{\theta} - \theta)^2 + \lambda_{\text{height}} (\hat{h} - h)^2
\end{align}
where $\lambda_{\text{angle}}$ and $\lambda_{\text{height}}$ are the weight factors. The final loss is a weighted sum of the above four loss functions,
{\small
\begin{align}
    L = \lambda_{\text{BEV}} L_{\text{BEV}} + \lambda_{\text{head}} L_{\text{head}} + \lambda_{\text{feet}} L_{\text{feet}} + L_{\text{pose}}
\end{align}
}
The loss weights are set to $\lambda_{\text{height}}=0.02, \lambda_{\text{angle}}=2.0$,  $\lambda_{\text{head}}=\lambda_{\text{feet}}=1.0$, and $\lambda_{\text{BEV}}=8.0$ in all experiments.

\begin{table*}[t]
    \centering
    \footnotesize
    \begin{tabular}{c|l|l|c|c|c|c}
    Type & Architecture & Pretrain & \multicolumn{1}{c|}{\begin{tabular}[c]{@{}c@{}}BEV Heatmap\\ \textit{MSE} $\times 10^{-7}$ $\downarrow$\end{tabular}} & \multicolumn{1}{c|}{\begin{tabular}[c]{@{}c@{}}Localization\\ \textit{Chamfer} $D \times 1$m / $D_n$\% $\downarrow$\end{tabular}} & \multicolumn{1}{c|}{\begin{tabular}[c]{@{}c@{}}Local risk map\\ \textit{IoU\%} $\uparrow$\end{tabular}} & \multicolumn{1}{c}{\begin{tabular}[c]{@{}c@{}}Global risk\\ \textit{MSE} $\times 10^{-4}$ $\downarrow$\end{tabular}} \\ \hline \hline
    \multirow{2}{*}{DET} & Mask R-CNN \cite{he2017mask} & COCO \cite{lin2014microsoft} & 3.89 & 2.26 / 75.3 & 46.63 & 43.35 \\
    & Faster R-CNN \cite{ren2015faster} & & 2.71 & 1.53 / 51.0 & 67.01 & 9.91 \\
    & CSP \cite{Liu_2019_CVPR} & CityPersons \cite{zhang2017citypersons} & 4.44 & 9.71 / 323.7 & 15.28 &  117.9 \\ \hline
    \multirow{4}{*}{CC} & CSRNet \cite{li2018csrnet} & NWPU-Crowd \cite{wang2020nwpu} & 337 & 4.13 / 137.7 & 17.11 &  $3.10 \times 10^4$ \\
     & DSSINet \cite{liu2019crowd} & ShanghaiTech B~\cite{Zhang_2016_CVPR} & 4.94 & 3.95 / 131.7 & 29.71 & 51.01 \\
     & IV-Net (Head) & CityUHK-X \cite{kang2017incorporating} & 5.36 & 3.90 / 130.0 & 30.01 & 40.19 \\
     & IV-Net (Feet) & & 8.25 & 3.65 / 121.7 & 22.41 & 9.03 \\
     & \textit{CC oracle} & --- & 5.61 & 3.51 / 117.0 & 30.17 & 48.13 \\ \hline

    \multirow{4}{*}{SDCA} & BEV-Net (Ours) & \multirow{4}{*}{CityUHK-X-BEV} & \textbf{1.34} & {1.25 / 41.7} & \textbf{71.25} & \textbf{6.24} \\
     & \quad - Feet only & & 1.38 & 1.26 / 42.0 & 68.12 & 7.62 \\
     & \quad - Head only & & 2.03 & 1.32 / 44.0 & 67.43 & 8.95 \\
     & \quad - No group transf. & & 1.36 & \textbf{1.24 / 41.3} & 69.65 & 7.08\\
    \end{tabular}
    \caption{SDCA performance, tested on CityUHK-X-BEV. Under BEV-Net: ``Head/feet only'' removes feet/head branch from model (fig.~\ref{fig:net_arch}); ``No group transf.'' uses a plain head branch w/o group transforms (sect.~\ref{sect:group_transform}). \textit{CC oracle} uses ground-truth head heatmap and camera pose.}
    \label{tab:main_results}
\end{table*}

\begin{figure*}[t]
    \centering
    \includegraphics[width=0.9\linewidth]{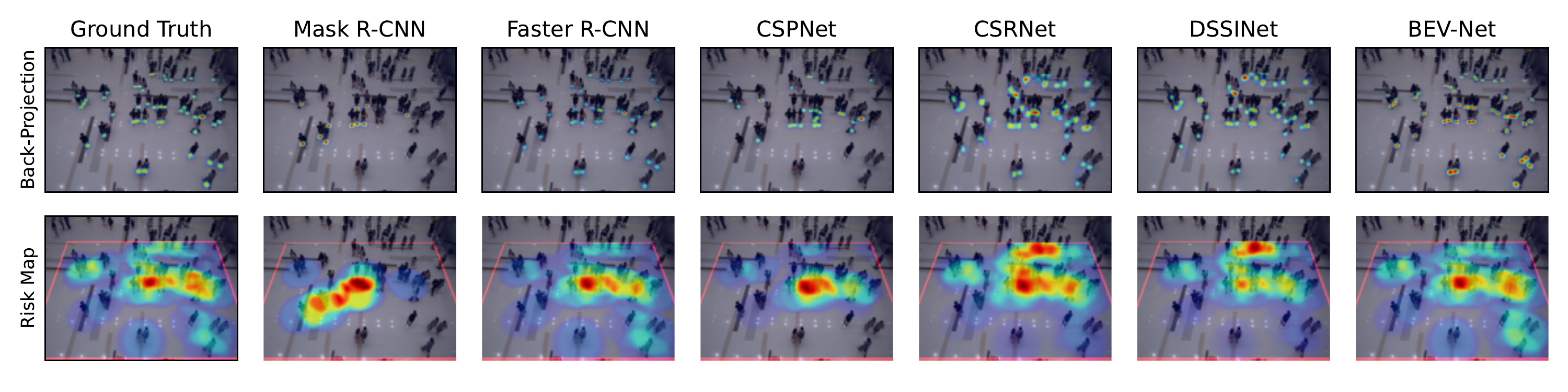}
    \caption{BEV localization \& risk heatmaps from different models. Both maps are back-projected to image space for visualization.}
    \label{fig:bev_maps}
\end{figure*}

\vspace{-0.5em}
\section{Experiments}
\vspace{-0.5em}

In this section we present experimental evaluations of SDCA on the CityUHK-X-BEV dataset.

\vspace{-0.5em}
\subsection{Experimental setup}
\vspace{-0.5em}

\noindent
{\bf Training procedure.}
The head, feet and pose branches are pretrained for 50 epochs before training the BEV-Net model. All the training process uses \textit{AdamW}~\cite{loshchilov2019decoupled} with learning rate $lr=0.0008$ exponentially decreasing by factor 0.98 per epoch. A batch size of 8 and a train-validation split of 4:1 was used in all experiments.
After pre-training, the BEV-Net is trained end-to-end. The BEV branch is first trained for 5 epochs with frozen pre-trained branches. All branches are then unfrozen step by step and jointly trained for 195 epochs.

\noindent
{\bf Comparisons.}
We compare BEV-Net to two types of baselines. A \emph{Detection}-based approach (DET) utilizes a person detector combined with the pretrained pose branch used by BEV-Net. The bottom center of each bounding box is used as the feet location $\rvq_i^j$. These locations are then converted to world coordinates in BEV using ${_I\mH_B}^{-1}\rvq_i^j$ with projection matrix ${_I\mH_B}$ estimated from the predicted camera pose (see section \ref{sect:gt_map}). We use pretrained CSPNet~\cite{Liu_2019_CVPR}, Mask R-CNN~\cite{he2017mask} and Faster R-CNN~\cite{ren2015faster}. The Faster R-CNN is finetuned on pseudo ground-truth bounding boxes generated from head/feet locations with aspect ratio $1 / (3\cos\theta)$.

A \emph{Counting}-based approach (CC) uses standard crowd-counting networks to generate IV head heatmaps, which we project to BEV using the BEV-Transform and the same pose branch used in DET methods. We use four networks: finetuned CSRNet~\cite{Liu_2019_CVPR} and  DSSINet~\cite{liu2019crowd}; and our IV-Net, which is the head/feet branch in BEV-Net. The vertical displacement between head locations and the ground plane is compensated for by subtracting an average pedestrian height used in \cite{kang2017incorporating}---1.75 meters---from the predicted camera height. To explore the upper bound of counting-based methods for SDCA, we introduce a \textit{CC oracle} which uses the ground-truth camera pose parameters and head maps.

Various ablations of the proposed BEV-Net are also evaluated. To study the effect of feet and head branches, we remove each of them from the architecture, leading to ``Head only'' and ``Feet only'' variants. We further ablate the head branch by replacing its group BEV-Transform module with a naive BEV homography (``No group transf.'').

\vspace{-0.5em}
\subsection{Quantitative results}
\label{sect:results}
\vspace{-0.5em}

\noindent
{\bf Model performance.} Table~\ref{tab:main_results} summarizes the performance of all methods in terms of the evaluation metrics of Section~\ref{sect:eval_protocol}. The BEV-Net outperforms all detection- and crowd counting-based methods by a significant margin.
Among different evaluation criteria, local and global risk estimates,and normalized Chamfer distance showed the most significant difference between models: BEV-Net  obtains  over $25\%$ higher IoU in local risk  prediction, $7\times$ lower error in global risk, and a $20\%$ reduction in normalized Chamfer Distance. While detection methods like Faster R-CNN can achieve relatively good localization performance with tight bounding boxes after finetuning, their risk estimates remain unreliable due to low recall from missed pedestrians. CC networks like DSSINet suffer from poor ground-plane localization, which hurts their SDCA performance in all criteria.
Notably, even \textit{CC oracle} with ground-truth head locations and camera pose or the IV-Net that predicts feet heatmap fails to meet the accuracy of the BEV-Net due to poor localization performance or feet occlusion.  This performance gap indicates that ground-plane modeling is essential for SDCA tasks, which cannot be effectively addressed by conventional detection or crowd counting approaches that operate solely in camera view.

\noindent
{\bf Ablation study.} Also reported in Table~\ref{tab:main_results} are ablated variants of BEV-Net. First, both ``Head only'' and ''Feet only'' models performed worse than the multi-branch BEV-Net. This suggests that both \emph{feet} and \emph{head} locations are important for SDCA, which is intuitive: Head locations do not lie on a shared 2D plane due to height variations among the crowd, making it challenging to estimate real-world distances; feet locations lie on the ground plane, but are often occluded in the camera view. Among the two, feet modeling is most effective. The ``Head only'' model struggles to produce accurate BEV maps and localization results. This is likely to be the case for all but very crowded scenes, with large amounts of feet occlusion. Second, the group BEV-Transform module of section~\ref{sect:group_transform} enabled considerable gain  over the vanilla implementation (``No group transf.'') under IoU of local risk prediction and global risk error. This confirms that modeling height variations within the population is beneficial for transforming image coordinates into BEV with high precision. Other ablation studies are presented in the supplemental material.

\begin{figure}[t]
    \centering
    \includegraphics[width=0.95\linewidth]{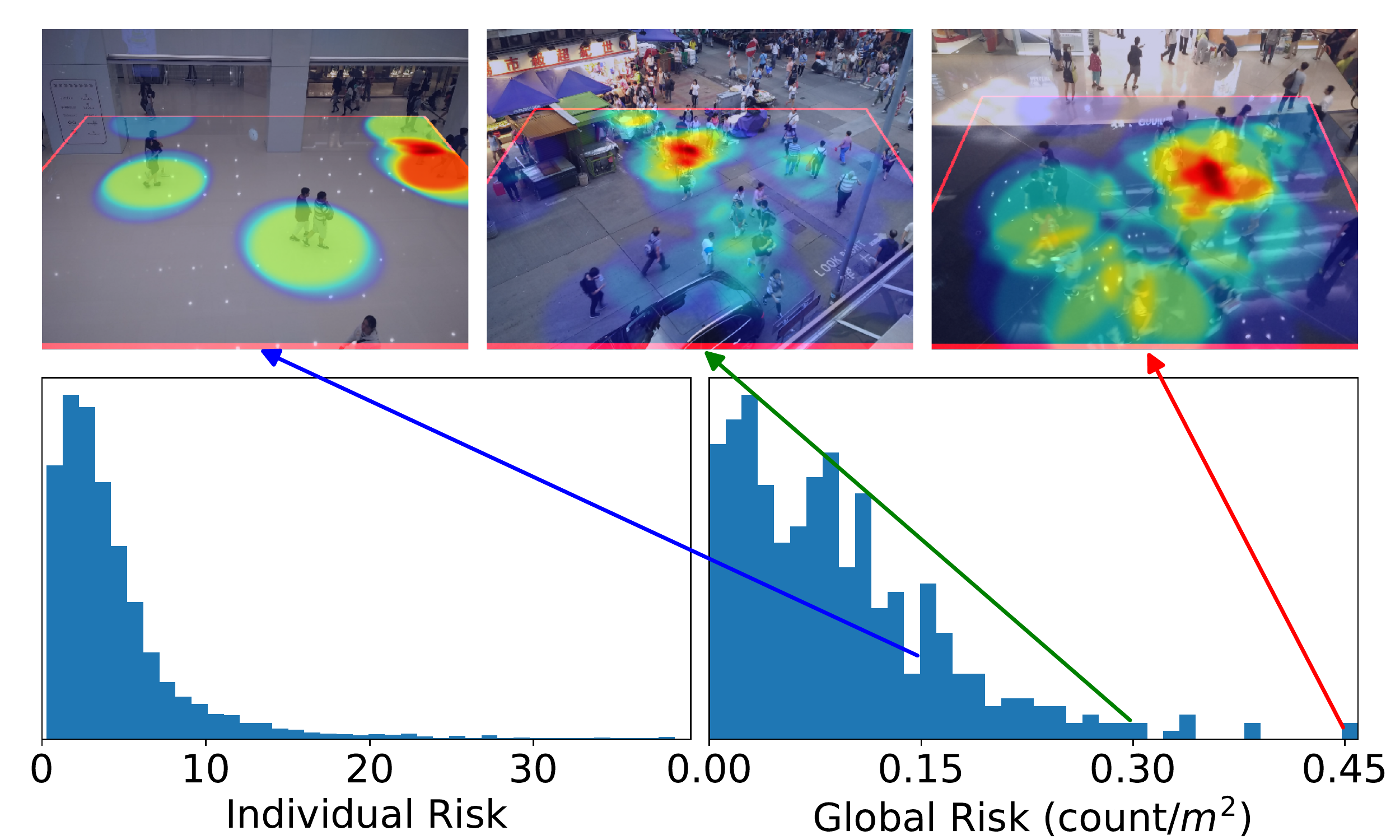}
    \caption{\textbf{Top}: Example scenes of different global risk level. \textbf{Bottom}: Histogram of individual and scene risks in test set.}
    \label{fig:top_risk}
\end{figure}

\subsection{Qualitative results}

We next evaluate the visual quality of the BEV-Net results, and discuss its possible applications.

\noindent
{\bf BEV heatmaps.} Figure~\ref{fig:bev_maps} compares BEV heatmaps and risk maps predicted by  different methods. Mask R-CNN~\cite{he2017mask}, Faster R-CNN~\cite{ren2015faster} and CSPNet~\cite{Liu_2019_CVPR} suffer from low recall of people detection, such that the risk maps fail to identify all regions with high risk of infection. CSRNet~\cite{li2018csrnet} and DSSINet~\cite{liu2019crowd} can capture more risky areas, but are unable to predict the risk level correctly due to the ambiguity in head heights. BEV-Net produces the closest localization and risk heatmaps to ground-truth. Note how it accurately predicts the risk ``hot-spots" inside clusters of people.

\noindent
{\bf Risk-based retrieval.} The multi-modal outputs from BEV-Net enable retrieval of images, individuals and clusters with highest risk of infection. Figure~\ref{fig:top_risk} shows the  distribution of individual and scene risks in the dataset and examples of events of different risk level. Individual risk is measured by the local risk level at the ground-plane location of each person. The graph shows that only 30\% of detected individuals are in compliance with the social distancing rule which prevents \emph{two or more people} from gathering together (\ie risk $\ge 2$). Under a less restrictive rule that relaxes the threshold to \emph{five people}, the compliance rises to 72\%. We believe this type of analytics is of interest for public health experts, e.g to  estimate transmission factors in real world scenes. Similarly, the global risk measure can be used to detect events of high risk, where viral transmission is most likely to occur.

\begin{figure}[t]
    \centering
    \includegraphics[width=0.95\linewidth]{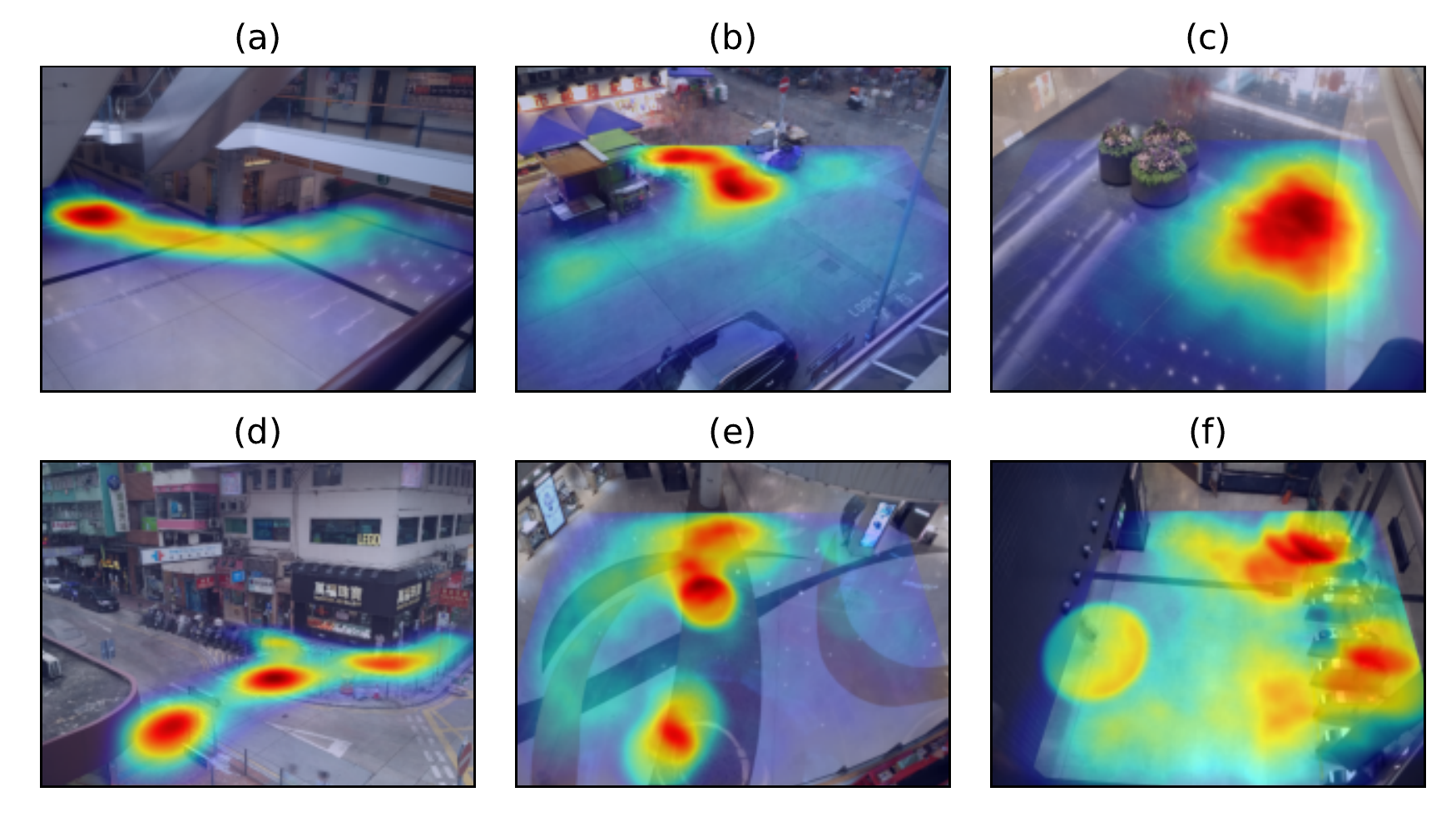}
    \caption{Mean risk map for test scenes, showing that people tend to gather at escalator entrances, near pedestrian crossings, vendor stalls, etc. Some results merit further investigation, e.g. why people prefer to use the gates near the two ends in (f).}
    \label{fig:scene_risk}
\end{figure}

\noindent
{\bf Scene risk analysis.} While we have so far focused on image measurements, BEV-Net can also be used to estimate intrinsic risk profiles of scenes. Figure~\ref{fig:scene_risk} shows the average risk map, over the test set, of several scenes. It can be observed that high-risk areas coincide with entrances, corners, passages or escalators.
These risk maps could be used by public health decision makers to identify potential infection hot-spots and place obstacles or warning signs in the scene to mitigate infection risks.

\vspace{-0.5em}
\section{Conclusion}
\vspace{-0.5em}

In this work, we have introduced the problem of social distancing compliance assessment on busy public spaces, from wide field-of-view cameras, without the need for manual camera calibration or  introduction of scene markers. A novel benchmark was proposed for this problem, where models are evaluated on their capability to localize people in the ground plane through geometric reasoning, and to identify regions where social distancing is violated. A multi-branch architecture, BEV-Net, was then presented, which fuses information from head and feet annotations to generate a BEV reconstruction of pedestrian locations. Experiments have shown that BEV-Net exceeds baseline methods under all evaluation metrics. Several applications of interest for public health decision makers have also been discussed.

\paragraph{Acknowledgements.} This work was partially funded by NSF awards IIS-1924937, IIS-2041009, NVIDIA GPU donations, and a gift from Amazon. We also acknowledge and thank the use of the Nautilus platform for some of the experiments discussed above.
ABC acknowledges support from a grant from the Research Grants Council of the Hong Kong Special Administrative Region, China (Project No. CityU 11212518).

{\small
\bibliographystyle{ieee_fullname}
\bibliography{ref}
}

\end{document}

%%%%%%%%% TITLE
\title{BEV-Net: Assessing Social Distancing Compliance \\ by Joint People Localization and Geometric Reasoning\\[5mm]Supplemental Material}

\author[1]{Zhirui Dai}
\author[1]{Yuepeng Jiang}
\author[1]{Yi Li}
\author[1,2]{Bo Liu}  % update Bo's affiliation
\author[3]{Antoni B. Chan}
\author[1]{Nuno Vasconcelos}
\affil[1]{Department of Electrical and Computer Engineering, UC San Diego}
\affil[2]{}
\affil[3]{Department of Computer Science, City University of Hong Kong}
\affil[ ]{\texttt{\small \{zhdai,yuj009,yil898,boliu\}@eng.ucsd.edu, abchan@cityu.edu.hk, nvasconcelos@ucsd.edu}}

\maketitle
\ificcvfinal\thispagestyle{empty}\fi

\appendix

%%%%%%%%% BODY TEXT

\section{Dataset Annotation}

\paragraph{Annotation procedure.}

The original CityUHK-X dataset \cite{kang2017incorporating} contained the head annotations of all people in the scene, as well as extrinsic camera parameters in the form of height $h$ and pitch angle $\theta$ relative to ground plane. The intrinsic parameters were assumed available at training and test time. As the height of each individual is unknown, head locations are not sufficient to recover pedestrians' locations in the world coordinates. Therefore, we used Amazon Mechanical Turk to annotate feet locations of each person, with one-to-one correspondence to the head locations.

\begin{figure}
    \centering
    \includegraphics[width=\linewidth]{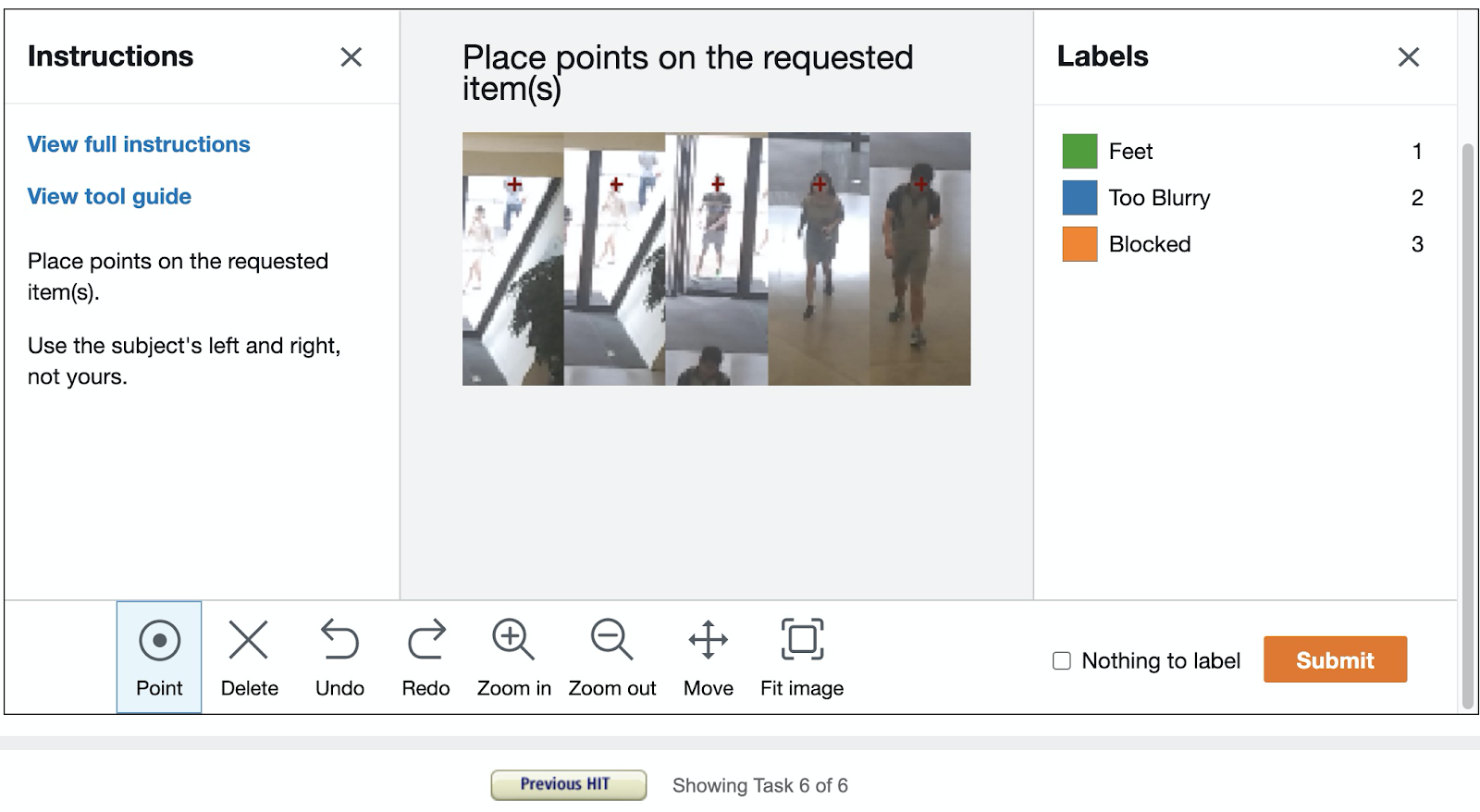}
    \caption{Annotation interface on MTurk.}
    \label{fig:mturk}
\end{figure}

As the number of people in each scene varies greatly (minimum 1 to maximum 121), the scene images are preprocessed into rectangular crops around each head location. The size of rectangles are selected adaptively to make sure that each crop contains the whole person selected in the original image. Given each crop with marked head location, workers are required to locate the midpoint between both feet that correspond to the same person (\figref{fig:mturk}); In crowded areas where one or both feet are occluded by objects or other pedestrians, workers are expected to provide their best estimate of feet location, or indicate that too little information is available to do so.

Each of the crops is assigned to three workers. The annotated coordinates from each worker are averaged after the exclusion of outliers. If at least two workers think they could see the feet clearly of the given person in the crop, then the crop is marked `valid' (clearly visible). Otherwise, the feet of the given person are marked to be occluded.

\paragraph{Annotation outcome.}

\begin{figure}
    \centering
    \includegraphics[width=0.48\linewidth]{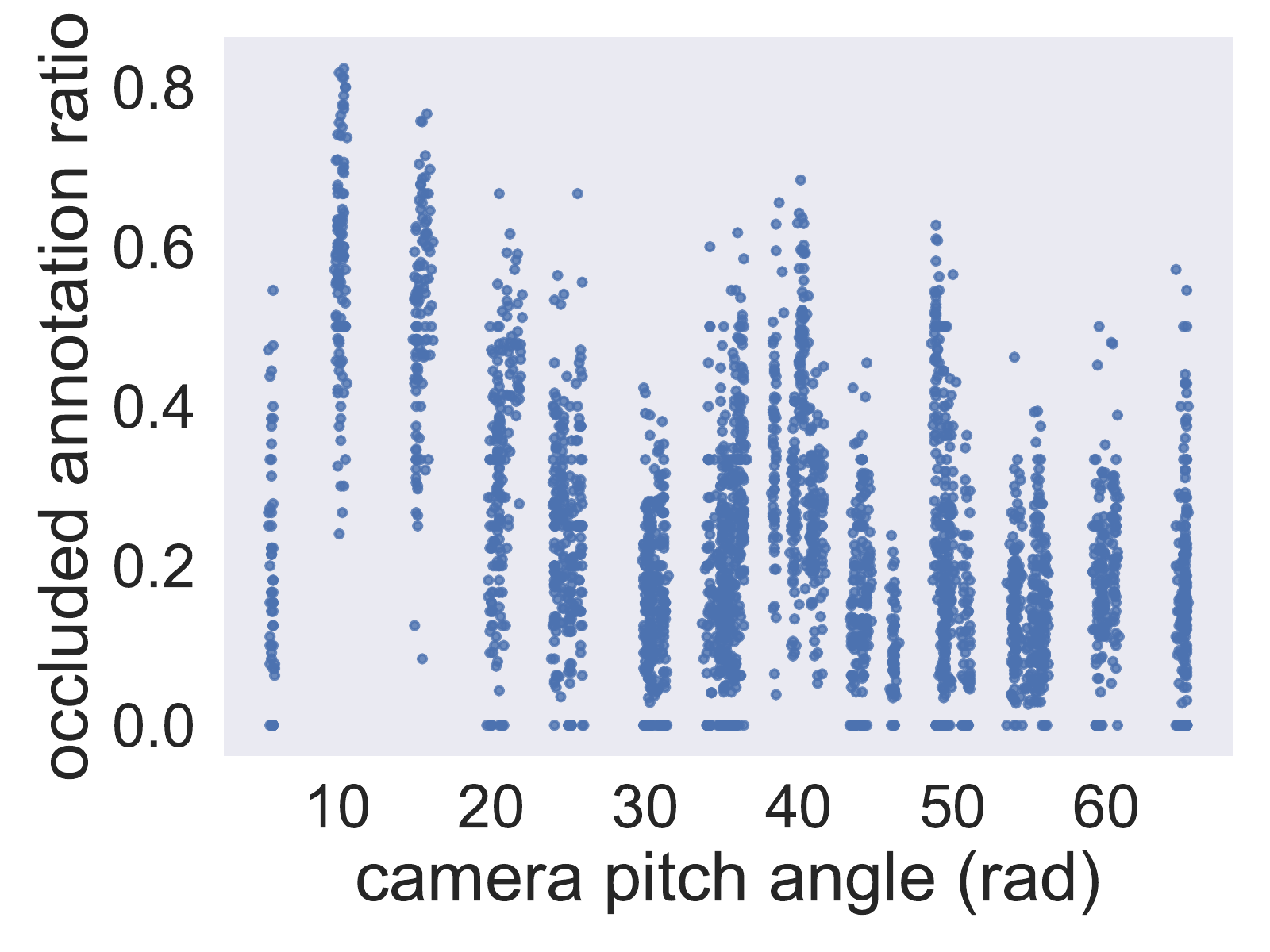} \hfill
    \includegraphics[width=0.48\linewidth]{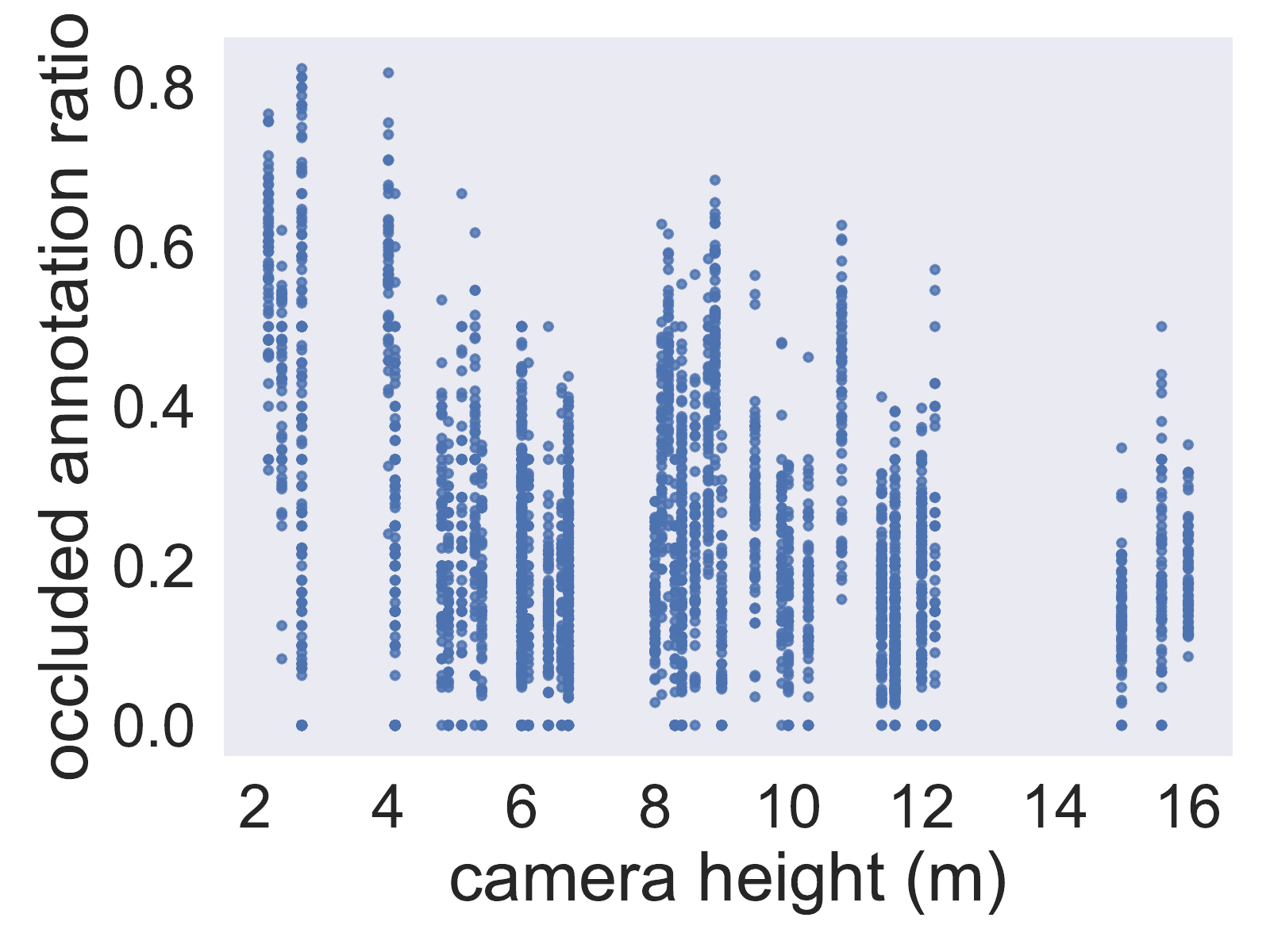}
    \caption{Percentage of estimated annotations from occluded body parts. More occlusion is found at smaller pitch angles and lower camera heights.}
    \label{fig:dataset_stats}
\end{figure}

87,746 feet locations were annotated using the procedure described above. Among them, 63,669 (72.56\%) were clearly visible and 24,077 (27.44\%) occluded. Figure \ref{fig:dataset_stats} shows the percentage of estimated annotations due to occluded body parts as functions of camera height and angle. The statistics reveal that occlusion occurs more frequently with low camera height and small pitch angles, making social distancing detection particularly challenging in these scenarios.

\section{Homography Derivation}

The setting of the camera is shown as figure {\color{blue} 2} in the main text. The origin of world coordinate is set to be the camera's perpendicular projection on the ground plane, and the yaw angle of camera is set to be 0 by aligning it with the $x$-axis of world coordinates. We further assume that the camera has zero roll angle, \ie its view is straightened to the horizon. This is a reasonable setting for most surveillance systems. Given the camera's height $h$ and pitch angle $\theta$, the transformation from the world frame to the optical frame, ${_O\rmT_W}$, is given by

\begin{align}
\begin{aligned}
    {_O\rmT_W} &= {_O\rmT_C}\ {_W\rmT_C}^{-1} \\
    &= \left[\begin{matrix}
        0 & -1 & 0 & 0 \\
        0 & 0 & -1 & 0 \\
        1 & 0 & 0 & 0 \\
        0 & 0 & 0 & 1
    \end{matrix}\right] \left[\begin{matrix}
        \cos{\theta} & 0 & \sin{\theta} & 0\\
        0 & 1 & 0 & 0 \\
        -\sin{\theta} & 0 & \cos{\theta} & h \\
        0 & 0 & 0 & 1
    \end{matrix}\right]^{-1} \\
    &= \left[\begin{matrix}
        0 & -1 & 0 & 0 \\
        -\sin{\theta} & 0 & -\cos{\theta} & h\cos{\theta} \\
        \cos{\theta} & 0 & -\sin{\theta} & h\sin{\theta} \\
        0 & 0 & 0 & 1
    \end{matrix}\right],
\end{aligned}
\end{align}
where ${_O\rmT_C}$ is the transformation from the camera frame to the optical frame, and ${_W\rmT_C}$ is from the camera frame to the world frame.

In CityUHK-X-BEV dataset, the camera focal lengths $(f_u, f_v)$ are given and for generality, we suppose there is no optical skew nor image center displacement. Hence, the intrinsic matrix is
\begin{align}
    \rmK = \left[\begin{matrix}
        f_u & 0 & u_c^I \\
        0 & f_v & v_c^I \\
        0 & 0 & 1
    \end{matrix}\right]. \label{eq:K}
\end{align}

Denoting with $\rmP$ the canonical projection matrix, transformation from point $(x, y, z)$ in the world frame to coordinates $(u, v)$ in the image frame is given by
\begin{align}
    {_I\rmT_W} &= \rmK\rmP\ {_O\rmT_W}\\
    \left[\begin{matrix}u & v & 1\end{matrix}\right]^\top &= {_I\rmT_W} \left[\begin{matrix}x & y & z & 1\end{matrix}\right]^\top. \label{eq:iTw}
\end{align}

For a plane at $z=h_0$, we can easily get the projection of points in the plane by using the camera's relative height $h' = h - h_0$. So, $z=0$ and \eqref{eq:iTw} becomes
\begin{align}
    \left[\begin{matrix}u & v & 1\end{matrix}\right]^\top = {_I\rmH_W} \left[\begin{matrix}x & y & 1\end{matrix}\right]^\top,
\end{align}
where
\begin{align}
    {_I\rmH_W} = \left[\begin{matrix}
    u_c^I\alpha & -f_u & u_c^I h' \beta \\
    v_c^I\alpha - f_v\beta & 0 & h'(f_v\alpha+v_c^I\beta) \\
    \alpha & 0 & h'\beta
    \end{matrix}\right],
    \label{eq:iHW}
\end{align}
$\alpha = \cos \theta$, $\beta = \sin \theta$, $(f_u, f_v)$ are the horizontal and vertical focal length of the camera respectively, and $(u^I_c, v^I_c)$ the image center.

Since the BEV map is under a certain scale as equation {\color{blue} 2} in the main text, the transformation between BEV map coordinates and the world frame is
\begin{align}
    {_W\rmT_B} = \left[\begin{matrix}
        0 & -s & x_c + sH/2 \\
        -s & 0 & y_c + sW/2 \\
        0 & 0 & 1
    \end{matrix}\right],
    \label{eq:wTB}
\end{align}
where $H, W$ are the height and width of the BEV map, and $\left(x_c, y_c\right)$ is the world coordinate of the image center on the ground plane. Matrices of \eqref{eq:iHW} and \eqref{eq:wTB} are combined in equation {\color{blue} 5} of main text to build the transform from image frame to BEV.

\section{Network Architecture}

\begin{figure*}
    \centering
    \includegraphics[width=0.75\linewidth]{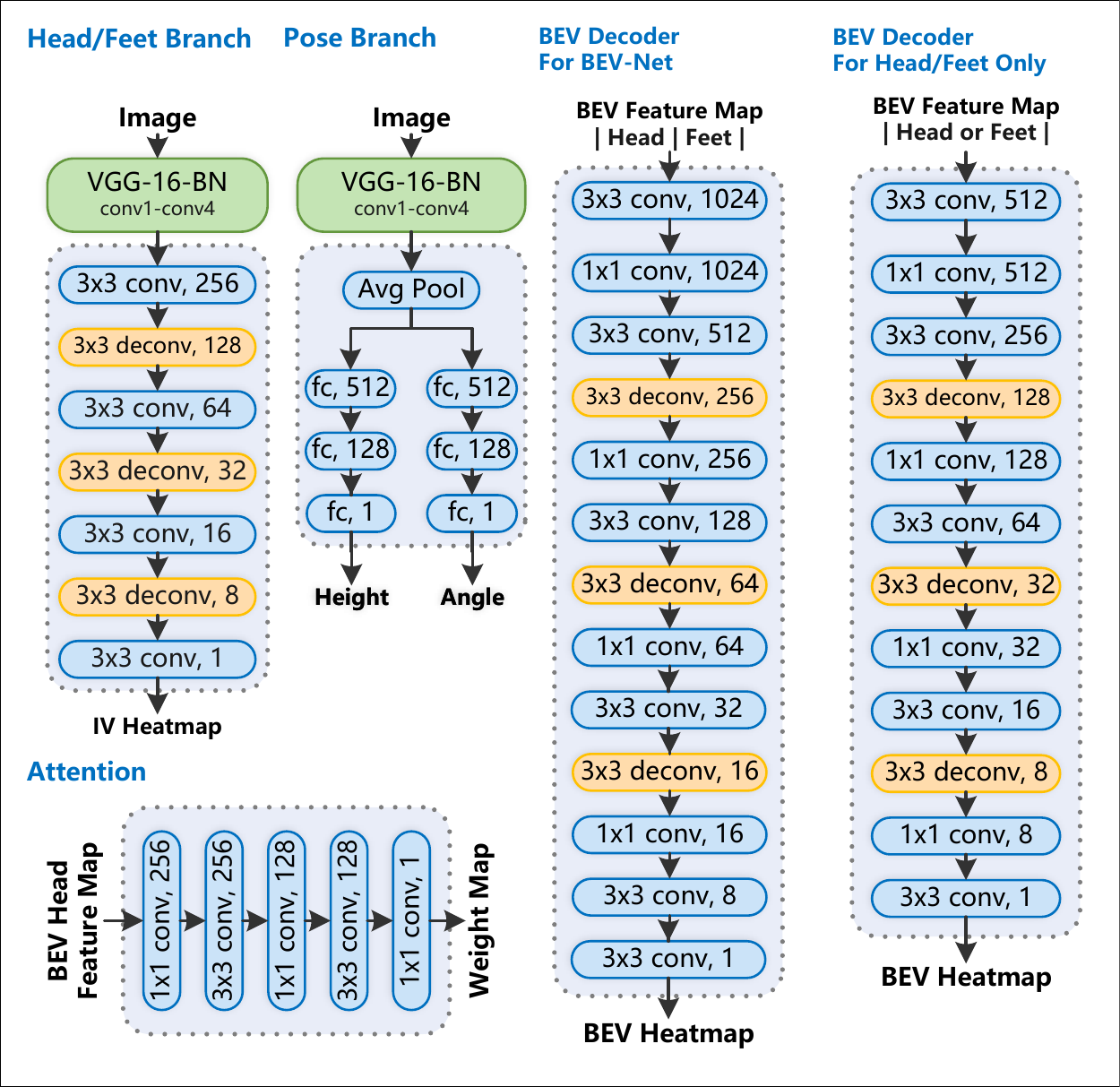}
    \caption{Network architectures. From left to right: IV (head/feet) branch, pose branch, BEV branch. The left bottom is the attention module. All \texttt{conv} layers have stride $s=1$; \texttt{deconv} layers use stride $s=2$. Nonlinearity, dropout~\cite{srivastava2014dropout} and batch normalization~\cite{ioffe2015batch} omitted between some layers for simplicity.}
    \label{fig:branch_arch}
\end{figure*}

Figure \ref{fig:branch_arch} summarizes the architecture for each branch of BEV-Net. Image-view (IV) branches estimate head or feet locations from input image using an encoder-decoder structure. The IV encoders followed the same design as the first 4 convolutional blocks of VGG-16 \cite{simonyan2014very} with batch normalization~\cite{ioffe2015batch}. Head and feet feature maps are then processed by a fully-convolutional decoder network into the IV heatmap. Pose branch uses fully connected layers stacked on top of a ResNet-101~\cite{he2016deep} feature extractor to regress camera height and pitch angle. The head and feet feature maps are projected into bird's eye view (BEV) using the BEV-Transform module (section {\color{blue} 4.2} of main text), then fed into the BEV decoder which predicts the final BEV heatmap.

\section{More Ablation Study}

\begin{figure}[t]
    \centering
\begin{minipage}{0.21\linewidth}
    \includegraphics[width=\linewidth]{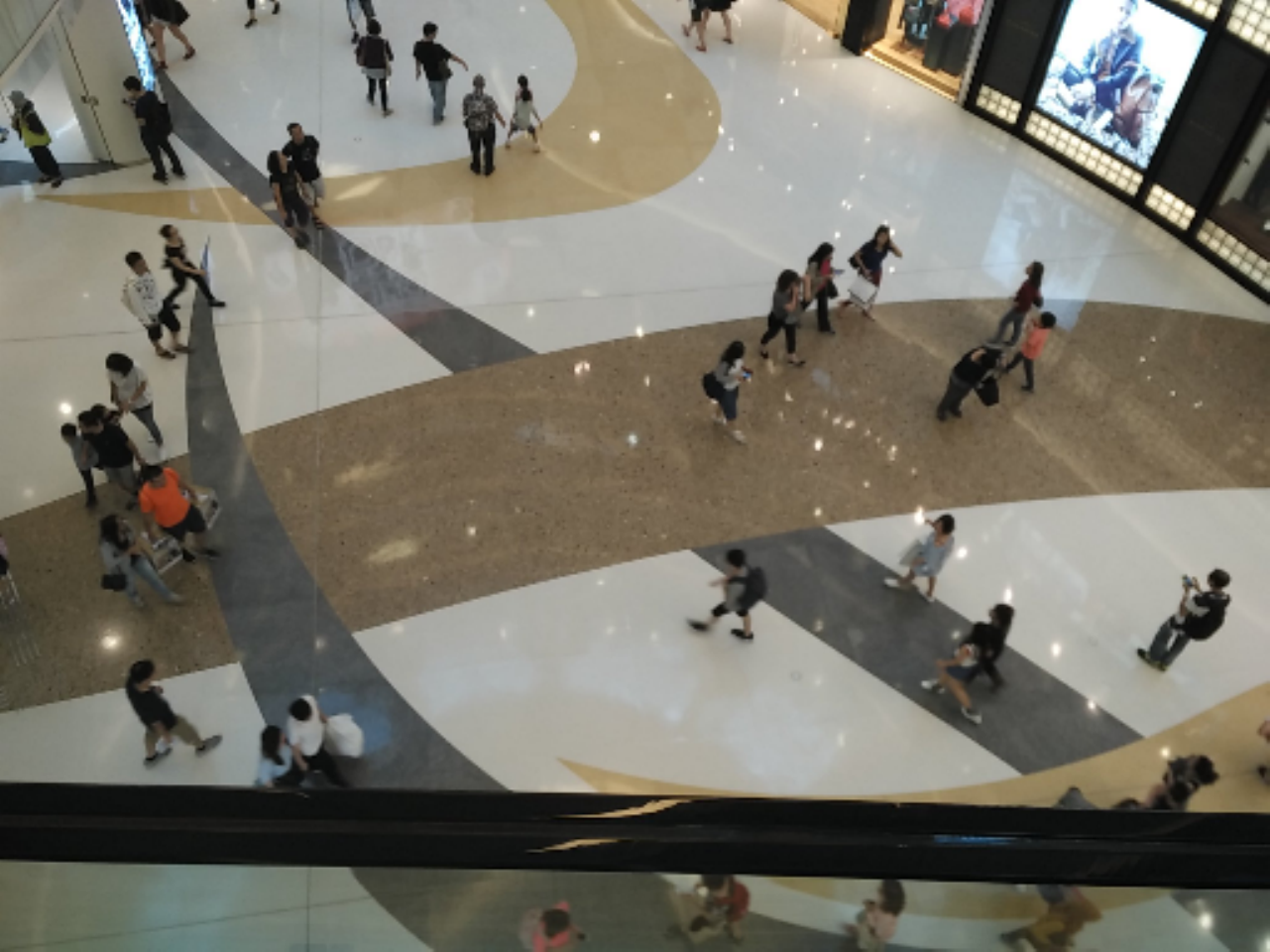}
    \caption*{$55.7^\circ, 12.0$m}
\end{minipage}\quad
\begin{minipage}{0.21\linewidth}
    \includegraphics[width=\linewidth]{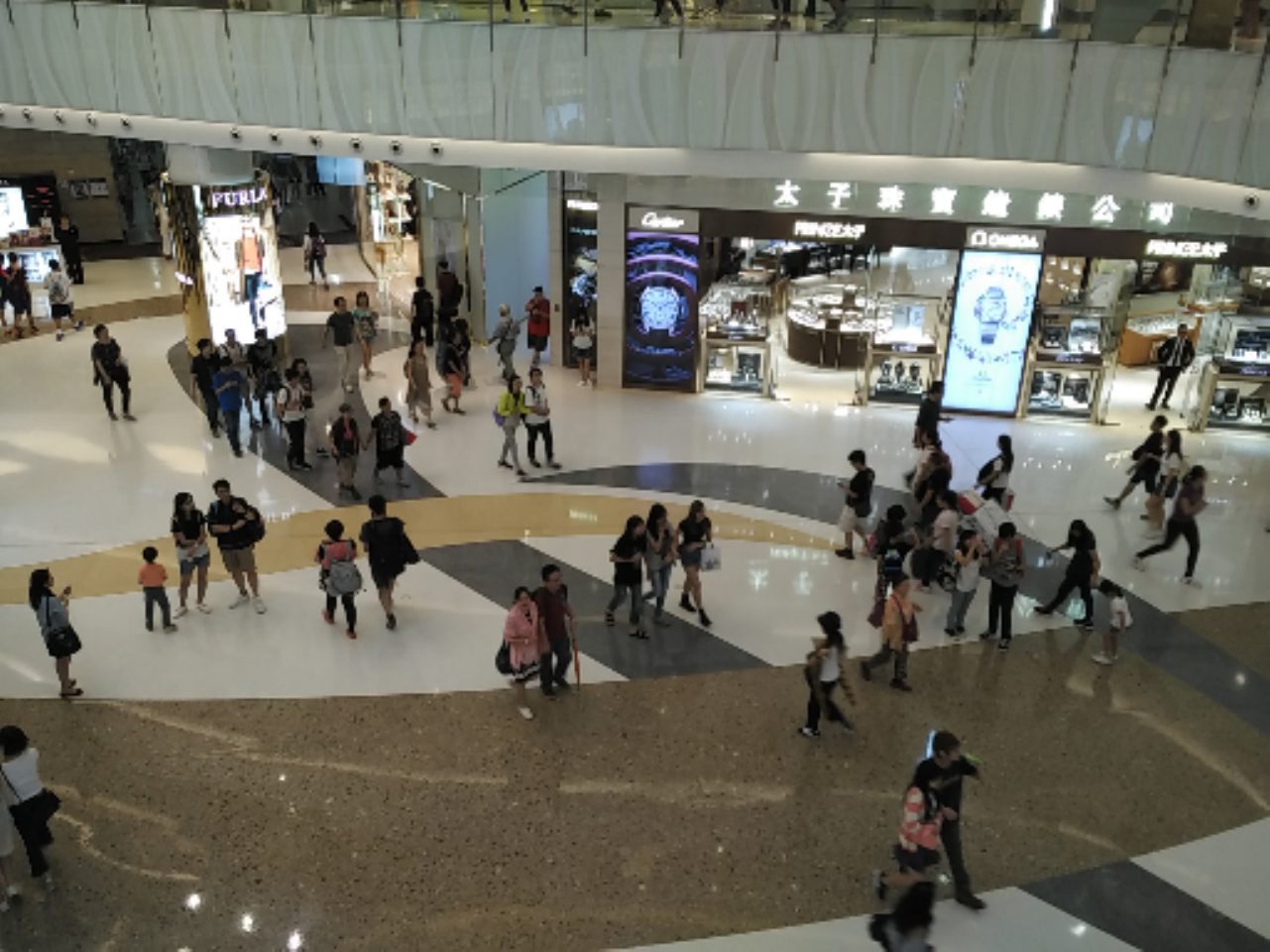}
    \caption*{$30.0^\circ, 8.0$m}
\end{minipage}\quad
\begin{minipage}{0.21\linewidth}
    \includegraphics[width=\linewidth]{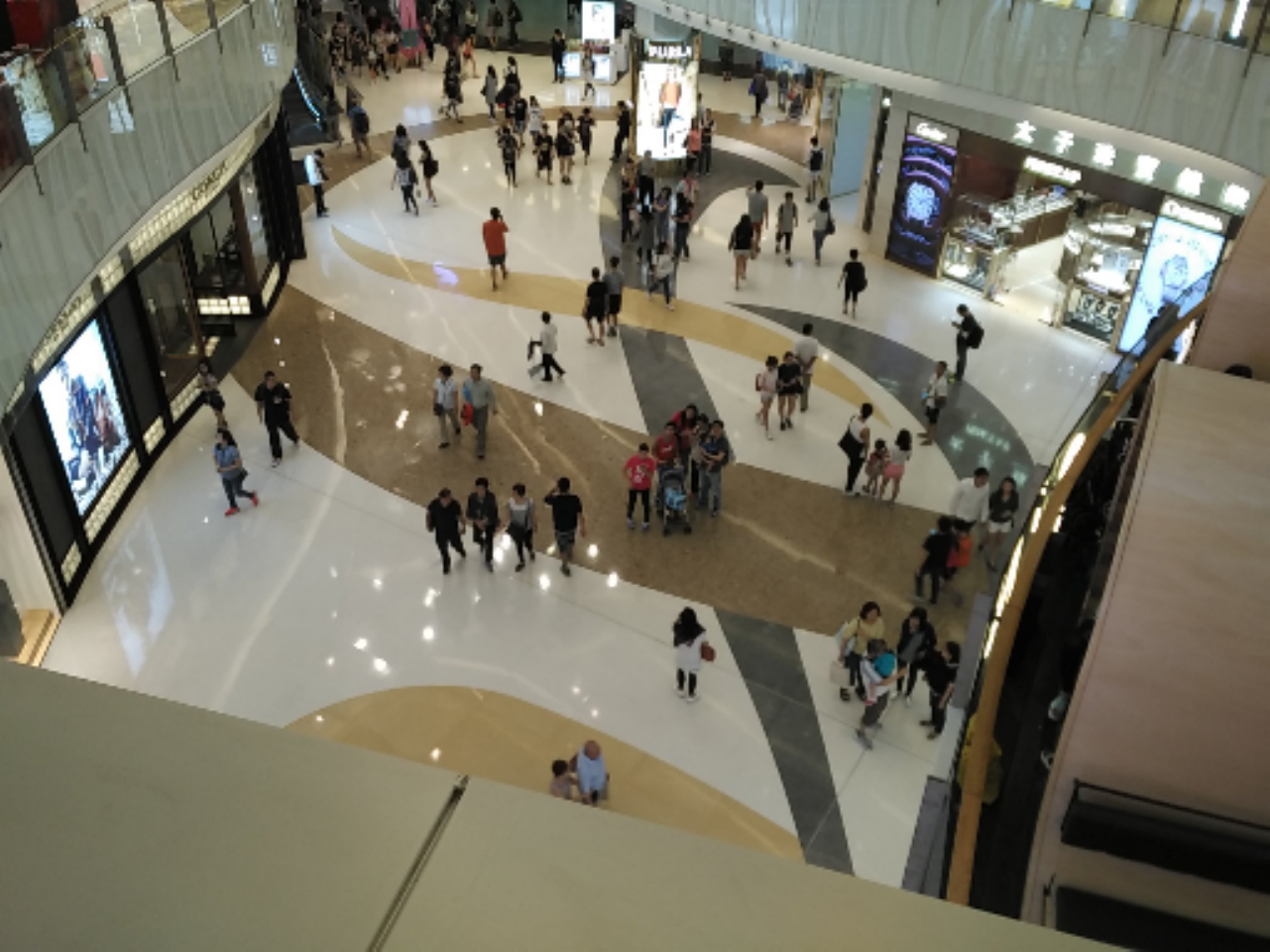}
    \caption*{$40.4^\circ, 12.0$m}
\end{minipage}
    \caption{Variation in camera poses in the same scene of CityUHK-X-BEV.}
    \label{fig:cam-pose-examples}
\end{figure}

\paragraph{Performance on split scene setting.}As shown in figure~\ref{fig:cam-pose-examples}, camera poses varies even within the same scenes of the CityUHK-X-BEV dataset. In the paper, we use the setting of PoseNet~\cite{posenet}, which trains and tests on the same scenes. We believe that this is the most suited for a public health setting, where there is usually some planing of the locations to monitor and data can be collected at those locations. In this setting, parameter variation is mostly due to camera motion (e.g. pan-zoom cameras), wind effects, etc. and usually less severe than even in figure~\ref{fig:cam-pose-examples}. A more drastic generalization to completely unseen scenes is a much more challenging task. We also test BEV-Net with some scenes unseen during training. The chamfer distance increases to 2.41/80.33\%, IoU of local risk drops to 54.86\%, and the global risk MSE is $50.14 \times 10^{-4}$. We can see that BEV-Net still outperforms most baselines.

\section{Qualitative Examples}

\begin{figure*}
    \centering
    \includegraphics[width=0.98\linewidth]{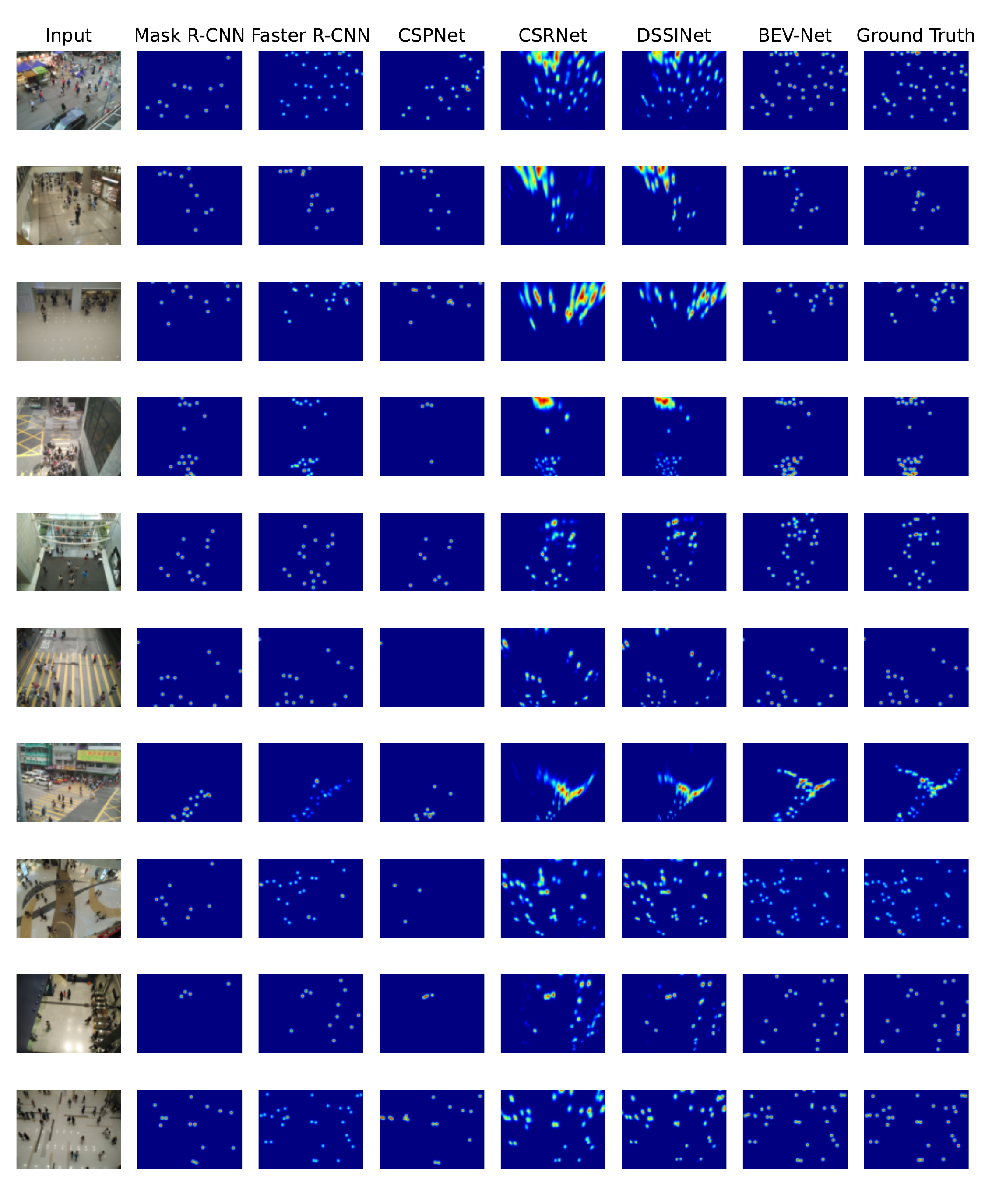}
    \caption{Qualitative comparison of \textbf{BEV heatmaps} between Mask R-CNN~\cite{he2017mask}, CSP~\cite{Liu_2019_CVPR}, CSRNet~\cite{li2018csrnet}, DSSINet~\cite{liu2019crowd} baselines and BEV-Net (ours). BEV-Net misses fewer people than detection methods~\cite{he2017mask,Liu_2019_CVPR} and produces more accurate localization than crowd counting approaches~\cite{li2018csrnet,liu2019crowd}.}
    \label{fig:qualitative_bevmap}
\end{figure*}

\begin{figure*}
    \centering
    \includegraphics[width=0.98\linewidth]{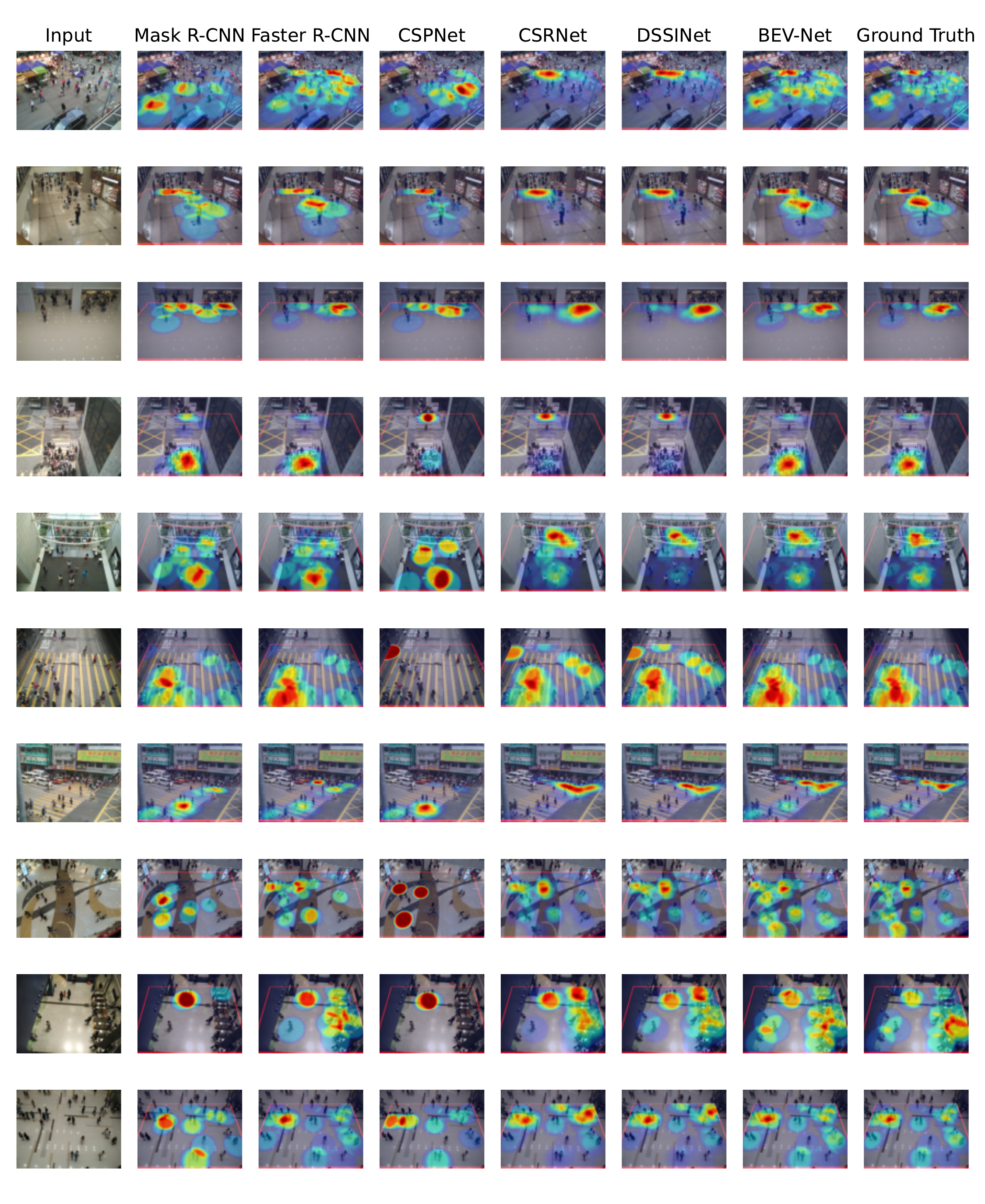}
    \caption{Qualitative comparison of \textbf{risk heatmaps} between Mask R-CNN~\cite{he2017mask}, CSP~\cite{Liu_2019_CVPR}, CSRNet~\cite{li2018csrnet}, DSSINet~\cite{liu2019crowd} baselines and BEV-Net (ours). Risk maps predicted by BEV-Net are closest to ground-truth.}
    \label{fig:qualitative_riskmap}
\end{figure*}

Figure \ref{fig:qualitative_bevmap} and \ref{fig:qualitative_riskmap} contain qualitative comparison of localization and risk predictions from the proposed BEV-Net and baseline approaches using detection~\cite{he2017mask,Liu_2019_CVPR} and crowd counting~\cite{li2018csrnet,liu2019crowd} backbones.
The results confirm the observations in main paper that detection methods have low recall for pedestrians far away, while counting methods fail to produce accurate localization in ground plane. In contrast, BEV-Net captures more people in crowded scenes, especially in areas far from the camera where occlusion is common, as well as those with extreme (close to 90$^{\circ}$) camera angles. This advantage translates to better localization and risk estimate performance, both in the visualizations and in quantitative results (table {\color{blue} 1} of main text).

{\small
\bibliographystyle{ieee_fullname}
\bibliography{ref}
}